\documentclass[10pt,twocolumn,letterpaper]{article}

\usepackage{style/wacv}              

\usepackage[numbers]{natbib}

\usepackage{xr}
\makeatletter

\newcommand*{\addFileDependency}[1]{
\typeout{(#1)}
\@addtofilelist{#1}
\IfFileExists{#1}{}{\typeout{No file #1.}}
}\makeatother

\newcommand*{\myexternaldocument}[1]{%
\externaldocument{#1}%
\addFileDependency{#1.tex}%
\addFileDependency{#1.aux}%
}

\myexternaldocument{supplementary}

\usepackage{graphicx}
\usepackage{amsmath}
\usepackage{amssymb}
\usepackage{amsthm}
\usepackage{booktabs}
\usepackage{multirow}
\usepackage{tablefootnote}
\usepackage{algorithm}
\usepackage{algpseudocode}
\usepackage{array}
\usepackage{balance}

\newcolumntype{?}{!{\vrule width 1pt}}

\newtheorem{theorem}{Theorem}

\usepackage[pagebackref,breaklinks,colorlinks]{hyperref}

\usepackage[capitalize]{cleveref}
\crefname{section}{Sec.}{Secs.}
\Crefname{section}{Section}{Sections}
\Crefname{table}{Table}{Tables}
\crefname{table}{Tab.}{Tabs.}

\def\wacvPaperID{65} 
\def\confName{WACV}
\def\confYear{2025}

\begin{document}

\title{SCAN-Edge: Finding MobileNet-speed Hybrid Networks for Diverse Edge Devices via Hardware-Aware Evolutionary Search}

\author{
Hung-Yueh Chiang, Diana Marculescu\\
The University of Texas at Austin\\
Chandra Family Department of Electrical and Computer Engineering\\
{\tt\small \{hungyueh.chiang, dianam\}@utexas.edu}
}
\maketitle

\begin{abstract}
Designing low-latency and high-efficiency hybrid networks for a variety of low-cost commodity edge devices is both costly and tedious, leading to the adoption of hardware-aware neural architecture search (NAS) for finding optimal architectures.
However, unifying NAS for a wide range of edge devices presents challenges due to the variety of hardware designs, supported operations, and compilation optimizations. 
Existing methods often fix the search space of architecture choices (\emph{e.g.,} activation, convolution, or self-attention) and estimate latency using hardware-agnostic proxies (\emph{e.g.,} FLOPs), which fail to achieve proclaimed latency across various edge devices. 
To address this issue, we propose \textbf{SCAN-Edge}, a unified NAS framework that jointly searches for \textbf{S}elf-attention, \textbf{C}onvolution, and \textbf{A}ctivatio\textbf{N} to accommodate the wide variety of edge devices, including CPU-, GPU-, and hardware accelerator-based systems.
To handle the large search space, \textbf{SCAN-Edge} relies on with a hardware-aware evolutionary algorithm that improves the quality of the search space to accelerate the sampling process. 
Experiments on large-scale datasets demonstrate that our hybrid networks match the \textbf{actual MobileNetV2 latency} for $224\times224$ input resolution on various commodity edge devices.
\end{abstract}

\section{Introduction} \label{sec:introduction}
Automatically designing deep learning architectures for specific hardware has become a compelling research topic. 
Both hardware vendors and software companies are developing accelerators to meet the operational budgets (e.g., hardware and energy costs) and serving constraints (e.g., latency and memory) of machine learning applications. 
Off-the-shelf models often fail to satisfy these constraints across a wide range of accelerators, necessitating the design of custom model architectures tailored to specific hardware platforms.
This design process is costly, as it requires machine learning experts to validate the performance of carefully tuned models on large-scale datasets and evaluate their efficiency on actual hardware.
Consequently, this issue significantly increases costs, prolongs the design cycle, and delays the deployment of machine learning services.

A significant body of prior work addresses this issue by proposing frameworks to automatically design optimal model architectures for different hardware or applications.
\citet{cai2020once} and \citet{yu2020bignas} train a one-shot supernet that searches for MobileNet-like subnets (convolution only) across various devices.
Recent approaches, such as those by \citet{gong2022nasvit} and \citet{tang2023elasticvit}, extend these methods to hybrid networks combining convolution and transformers. 
However, these methods often fix the search space of architectural choices (\emph{e.g.,} GELU activation, depth-wise convolution, or self-attention layers) for network stages.
This can be suboptimal for certain edge devices, as the optimal search space largely depends on both hardware implementation and compiler optimization.
For instance, a memory-bound operator like depth-wise convolution may not be efficient or optimal for a highly parallelized device \cite{lu2021optimizing, zhang2020high}. 
Additionally, prior works like \citet{gong2022nasvit} estimate model complexity using zero-cost proxies \cite{white2023neural} such as floating-point operations (FLOPs) and the number of parameters, which do not accurately reflect the \emph{actual latency} on target edge devices.
To illustrate this point, Figure \ref{fig:preliminary} and Section \ref{sec:preliminary} demonstrate that different edge devices favor different operations when actual hardware metrics are considered.

%
To address this issue, we propose a unified NAS framework, \textbf{SCAN-Edge}, which utilizes a weight-sharing supernet to search for a variety of hybrid networks optimized for edge devices. 
Our search space includes unified feedforward networks (unified FFNs, memory-bound), fused feedforward networks (fused FFNs, compute-bound), and multi-head self-attention layers (MHSAs), along with two activation functions, GELU and ReLU, to support a wide variety of commodity devices. 
During the search process, we incorporate calibrated latency lookup tables (LUTs) profiled on target devices and a learning-based accuracy predictor. 
These LUTs accurately estimate the end-to-end latency of the subnets. 
To find the optimal subnet from this huge search space, our search algorithm optimizes the search space based on the target hardware.
Our experiments demonstrate that hybrid networks discovered by SCAN-Edge match the actual \emph{MobileNetV2 latency} while providing \emph{ better accuracy} across a wide range of edge devices compared to prior approaches.

Our contributions are as follows:
\begin{itemize}
  \item We perform a comprehensive analysis of various hardware platforms focusing on self-attention, convolution, and activation operations.
  \item We created a unified framework that optimizes convolution and activation types, and the number and placement of MHSA layers for subnets that meet the latency constraints on a variety of devices.
  \item To handle the large search space, we introduce a hardware- and compiler-aware evolutionary algorithm that enhances search quality and speeds up the sampling process.
\end{itemize}
\section{Related Work} \label{sec:related_work}

\paragraph{Efficient vision transformers.}
Though \citet{dosovitskiy2020image}, \citet{touvron2021training}, and \citet{liu2021swin} show the competitive performance of vision transformers (ViTs) against convolutional neural networks (CNNs), deploying ViTs on edge devices is challenging due to the $O(n^2)$ complexity of attention layers.
Many works \cite{li2023rethinking, li2022efficientformer, pan2022edgevits, chen2021crossvit, chen2022mobile, maaz2022edgenext, mehta2021mobilevit} address this issue and propose lightweight attention operators.
\citet{li2022efficientformer} replace self-attention with pooling layers for early-stage high-resolution inputs.
\citet{li2023rethinking} and \citet{pan2022edgevits} downsample the attention resolution in exchange for model efficiency.
However, the lightweight operators with fewer floating-point operations (FLOPS) are not necessarily friendly to edge devices.
For instance, MobileViT \cite{mehta2021mobilevit} includes \emph{fold} and \emph{unfold} mobile-unfriendly operations to reorder the data memory before processing, which incurs additional latency during inference.
We design our supernet with a search space based on EfficientFormerV2 \cite{li2023rethinking} with MobilenetV2-like  \cite{sandler2018mobilenetv2} feedforward layers (FFNs) and global-local attention layers, which are friendly to hardware platforms.

\paragraph{Neural architecture search.}
NAS has achieved huge success in CNNs \cite{yu2020bignas, cai2020once, stamoulis2019single}.
Recent approaches \cite{chen2021searching, chen2021autoformer, liu2022uninet, su2022vitas} apply NAS to automatically design and improve ViTs architectures.
\citet{chen2021autoformer} propose a weight-entangled ViT supernet where all subnets can be trained at once and without further fine-tuning before deployment.    
\citet{chen2021searching} use a high-quality search space before identifying the ViTs.
However, the patch embedding layer with large kernel sizes and non-overlapping stride sizes are ill-supported by edge devices \cite{li2022efficientformer}.  
Hybrid networks (CNN-ViT) \cite{graham2021levit, li2022efficientformer} replace the patch embedding layer with a sequence of CNN stages to reduce feature resolution before passing to transformer stages.
Some works \cite{gong2022nasvit, tang2023elasticvit} focus on searching for efficient hybrid networks for mobile devices.
However, \citet{gong2022nasvit} only use FLOPS as the complexity metric during the search process, which does not truly reflect on-device latency.
%
%
Instead of using FLOPS, we build block-wise latency lookup tables (LUTs) for a wide variety of devices and calibrate the LUT estimations by linear regression from 10 subnet latencies obtained from real measurements.
Therefore, our method generalizes to and searches for optimal architectures across various commodity edge devices, characterized by actual device latencies.
\section{Preliminary Study} \label{sec:preliminary}

\begin{figure}[t]
    \centering
    \subfloat[\centering The first stage]{{\includegraphics[width=0.5\columnwidth]{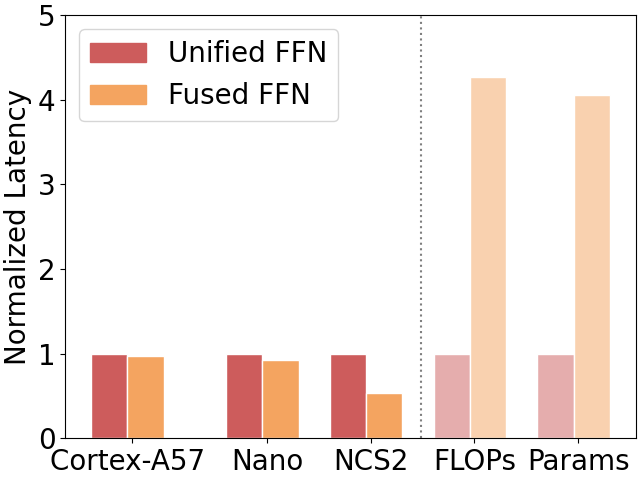} }}%
    \subfloat[\centering The last stage]{{\includegraphics[width=0.5\columnwidth]{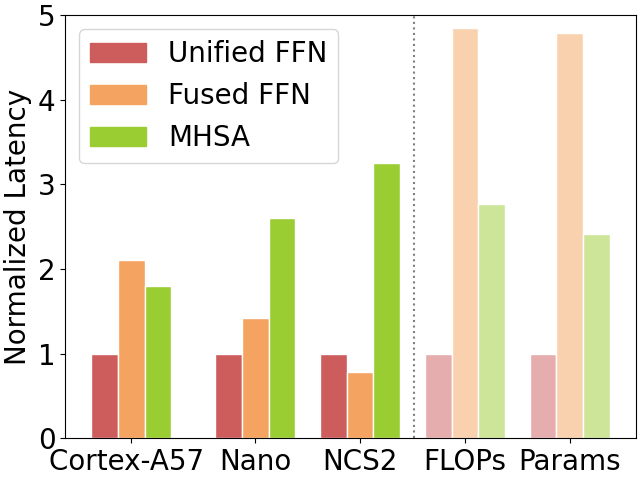} }}%
    \caption{
    We profile the latency and zero-cost proxies of EfficientFormerV2 S0 on different devices.
    (a) shows the latency of the first stage (FFNs only) with input size (h, w, c)=(56, 56, 32). While unified FFN has fewer FLOPS, it is bounded by memory and has a similar latency to fused FFN.
    (b) shows the latency of the last stage (FFNs and MHSAs) with input size (7, 7, 176). The stage latency is device-dependent and highly different from the proxies.
    }%
    \label{fig:preliminary}%
\end{figure}

In our preliminary study, we use EfficientFormerV2 S0 \cite{li2023rethinking} as our base architecture and experiment with different FFNs, activations, and self-attention ratios (expansion of $V$ dimension).
%
%
%
We follow EfficientFormerV2 and use the term, feedforward network, for convolution layers.

\paragraph{Zero-cost proxies \emph{vs.} Latency profiling.}
Zero-cost proxies such as the number of parameters and floating-point operations (FLOPS) are widely used in NAS for estimating the model complexity and latency due to their immediate availability.
However, FLOPS do not accurately capture the actual edge device latency.
In Figure \ref{fig:preliminary}, we show that estimating model complexity with zero-cost proxies fails to generalize to a wide range of edge devices.
For example, fused FFNs, despite having more FLOPS and parameters, have lower latency than MHSA layers on Nano and NCS2.
%


\begin{table}[t]
\scriptsize
\centering
\caption{We evaluate the accuracy (Acc.) on ImageNet 1k with different variants of  EfficientFormerV2 S0 and show end-to-end latency (ms) on different devices in the last column.}
\begin{tabular}{@{}ccc|c|ccc@{}}
\toprule
FFN     & Act. & $V$ ratio  & Acc. (\%) & Cortext & Nano TRT & NCS2 \\ \midrule \midrule
Unified & GELU & 4          & 75.7      & 190.3 & 21.4 & 47.0   \\ \midrule
Unified & GELU & \textbf{2} & 73.3      & 168.7 & \textbf{18.1} & 42.1 \\ \midrule
Unified & \textbf{ReLU} & 4 & 74.4      & \textbf{105.6} & 18.3 & 30.6    \\ \midrule
\textbf{Fused}  & GELU & 4  & \textbf{77.1}      & 237.6 & 26.2 & 36.6  \\ \midrule
\textbf{Fused}  & \textbf{ReLU} & 4 & 76.6         & 187.7 & 22.3 & \textbf{26.6}   \\ \bottomrule
\end{tabular}
\label{tab:preliminary}
\end{table}

\begin{figure}[t]
\centering
\includegraphics[width=0.8\columnwidth]{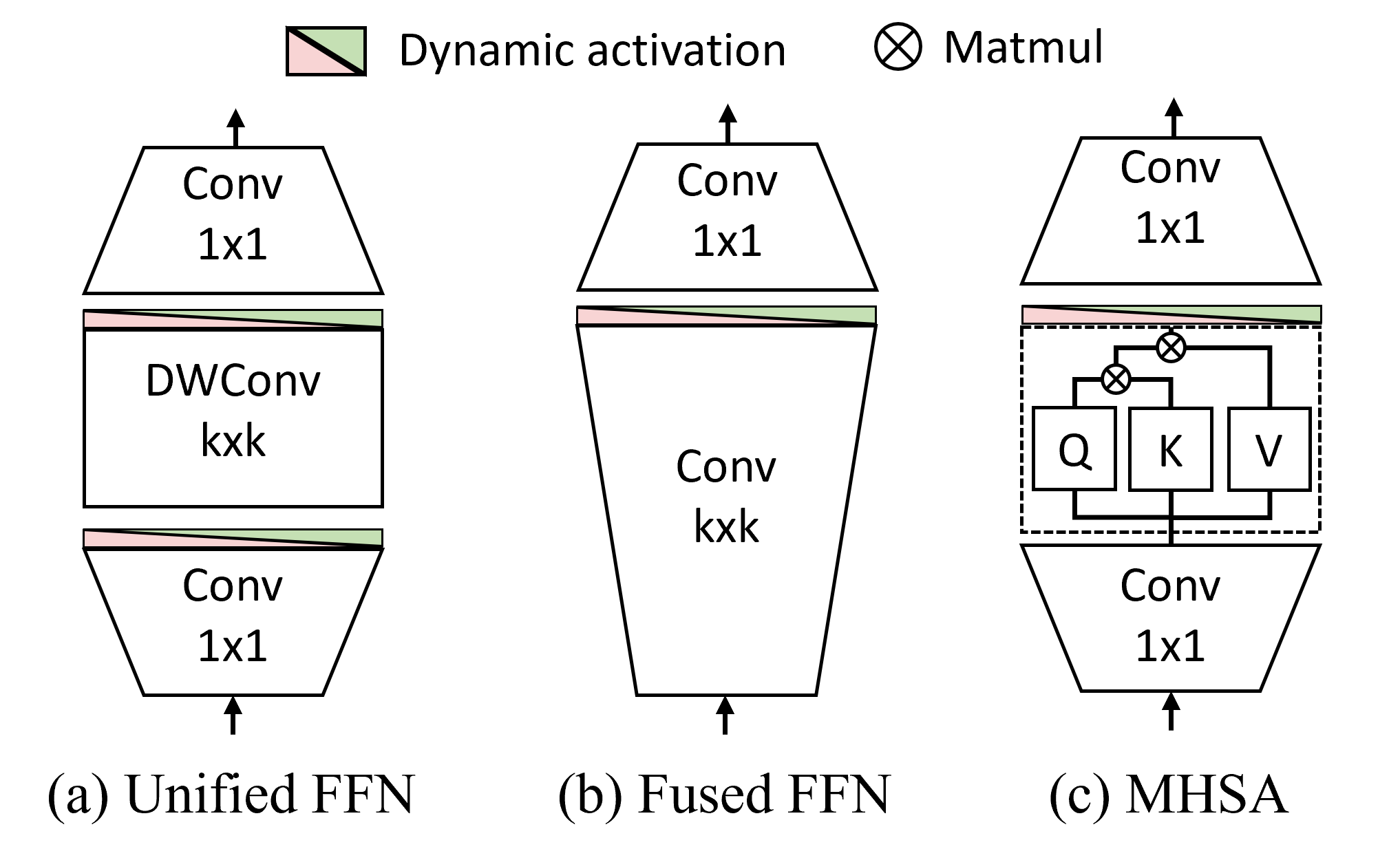}
\caption{We use three components in our supernet with dynamic activation layers. Each MHSA block is followed by either a unified FFN or a fused FFN. The components, \emph{e.g.,} residual connections, are simplified to avoid cluttering the figure.}
\label{fig:components}
\end{figure}

\paragraph{Unified FFN \emph{vs.} Fused FFN.}
Although unified FFNs learn spatial relationships in the feature maps by a depthwise convolution with fewer parameters and fewer theoretical FLOPS, they are highly memory-bound with low arithmetic intensity.
In Table \ref{tab:preliminary} and Figure \ref{fig:preliminary}, fused FFNs with vanilla convolutions improve accuracy by 1.4\% are not only well-supported by NCS2 but also better suited for early network stages among all devices.
We include two operators in our supernet to support more subnet choices for different edge devices, as shown in Figure \ref{fig:components}.

\paragraph{Feedforward network \emph{vs.} Multi-head self-attention layer.}
MHSAs learn the long-range dependencies in the feature map which greatly boosts performance.
However, they are the major bottleneck in latency (\emph{cf.} Figure \ref{fig:preliminary}).
As shown in Table \ref{tab:preliminary}, reducing the value ratio from $4$ to $2$ in the self-attention layers ($\texttt{dim}(V) = \texttt{dim}(\texttt{Input}) \times \texttt{ratio}$) hurts the accuracy by $2.4\%$.
Our framework searches the \emph{number} and the \emph{placement} of MHSA layers along with their expansion ratios for the value matrices.
For devices where MHSA is less suitable, our framework minimizes the number of MHSA layers within the architecture.
Conversely, if a compiler or implementation enhances MHSA support on a device, as seen in \cite{dao2022flashattention, dao2023flashattention2}, our search algorithm adaptively increases the number of MHSA layers to enhance accuracy while maintaining the same latency constraints.

\paragraph{GELU \emph{vs.} ReLU.}
While GELU activation \cite{hendrycks2016gaussian} improves accuracy, it is often not well supported by low-cost edge devices.
In contrast, most devices and compilers support conv-relu fusion which provides optimal latency \cite{onnx, openvino}.
As shown in Table \ref{tab:preliminary}, the latency is greatly improved by 3.1ms ($14.4\%$) when replacing GELU with ReLU \cite{agarap2018deep} in the EfficientFormerv2 S0.
We include GELU and ReLU as the activation choices for different layers in our supernet.

We conclude the preliminary study, and based on it, propose a unified NAS framework, named \textbf{SCAN-Edge}, for general edge devices.
Our search space includes device-friendly operators: two feedforward networks, the number and the placement of self-attention layers, and two activation functions. 
%

\section{One-shot Supernet}\label{sec:oneshot_supernet}

\begin{table*}[]
\scriptsize
\centering
\caption{Full search space. Our search space includes fused FFN (Fused), Unified FFN (unified), and Multi-head self-attention (MHSA). The third embedding layer (Embed 3) is an MHSA down-sampling layer (DS). We search the number of blocks in each stage ($N_{\text{ffn}}$ and $N_{\text{mhsa}}$) as well as the activation (GELU (G) or ReLU (R)), expansion ration $E^j_i$, and kernel size $K^j_i$ in each block. We abbreviate output resolution (Out Res.), and output channel width (Out Ch.). The channel width is represented in (min, max, step size). }
\label{tab:search_space}
\begin{tabular}{@{}c|cc|ccccc|c@{}}
\toprule
Stage                      & Type              & Out Res. & Out Ch. ($C_i$)               & Exp. ($E^j_i$)  & K size ($K^j_i$) & Act.  & \#choices   & Depth    \\ \midrule \midrule
Stem Conv                  & Conv              & 1/4      & $C_1$ = (24,  36, 4)               & -           & 3           & \{G, R\}   &  8   & -           \\ \midrule
Stage 1                    & \{Fused, Unified\} & 1/4      & $C_1$                  & \{2, 3, 4\} & \{3, 5\}    & \{G, R\}   &   96 & $N_{\text{ffn}}$ = \{2, 3\}  \\ \midrule
Embed 1                    & Conv              & 1/8      & $C_2$ = (40,  64, 8)   & -           & 3           & -        &   16 & -                               \\ \midrule
Stage 2                    & \{Fused, Unfied\} & 1/8      & $C_2$                  & \{2, 3, 4\} & \{3, 5\}    & \{G, R\}  &   96 & $N_{\text{ffn}}$ = \{2, 3\}    \\ \midrule
Embed 2                    & Conv              & 1/16     & $C_3$ = (96,  132, 12) & -           & 3           & -        &   16 & -                            \\ \midrule
\multirow{2}{*}{Stage 3}   & MHSA              & 1/16     & $C_3$                  & \{2, 3, 4\} & -           & \{G, R\} &   24 & $N_{\text{mhsa}}$ = $\{n | 0 \leq n \leq N_{\text{ffn}}\}$             \\ \cmidrule(l){2-9} 
                           & \{Fused, Unified\} & 1/16     & $C_3$                  & \{2, 3, 4\} & \{3, 5\}    & \{G, R\} &    96 & $N_{\text{ffn}}$ = \{6, 7, 8, 9\}   \\ \midrule
Embed 3                    & MHSA DS           & 1/32     & $C_4$ = (176, 248, 24) & \{2, 3, 4\} & -           & \{G, R\} &  96 &  -                                     \\ \midrule
\multirow{2}{*}{Stage 4}   & MHSA              & 1/32     & $C_4$                  & \{2, 3, 4\} & -           & \{G, R\}  &  24 & $N_{\text{mhsa}}$ = $\{n | 0 \leq n \leq N_{\text{ffn}}\}$    \\ \cmidrule(l){2-9} 
                           & \{Fused, Unified\} & 1/32     & $C_4$                  & \{2, 3, 4\} & \{3, 5\}    & \{G, R\} &  96  & $N_{\text{ffn}}$ = \{4, 5, 6\}        \\ \bottomrule
\end{tabular}
\end{table*}

\paragraph{Search space.}
We construct our supernet based on the EfficientFormerV2 backbone.
The supernet consists of four stages and three embedding layers.
The embedding layers adjust the model width, \emph{i.e.}, the channel dimension $C_i$, for the next stage.
Each block in the first two stages, $S_1$ and $S_2$, only consists of FFNs.
Every block in the last two stages, $S_3$ and $S_4$, consists of a MHSA followed by a FFN.
Our search algorithm determines the width $C_i$, the number of FFN blocks $N_{\text{ffn}}$ in every stage $S_i$, the number of MHSA $N_{\text{mhsa}}$ in the last two stages for the network, and the expansion ratios $E^{j}_{i}$ (FFN and MHSA) and kernel size $K^{j}_{i}$ (FFN only) for every block $j$ in stage $i$. 
The full search space is shown in Table \ref{tab:search_space}.

\paragraph{Dual feedforward network.}
%
We design our supernet with dual FFN to best accommodate various edge devices.
As shown in Figure \ref{fig:components}, in each FFN block, we provide two choices for searching: unified FFN and fused FFN with different kernel sizes (\emph{e.g.,} 3, 5) and expansion ratios (\emph{e.g.,} 2, 3, 4).
%
%
Only one set of weight matrices will be activated (unified or fused) in a block during training.
We empirically find dual FFNs converge slightly better than entangled FFNs (\emph{cf.} Suppl. \ref{dual_vs_entangled_ffn}, and Figure \ref{fig:entangled_vs_dual}). 
During training, we sample $N_{\text{ffn}}$ blocks for every stage $S_i$ and randomly switch between unified FFN and fused FFN.
For each FFN block, we sample the kernel size $K^{j}_{i}$, and expansion ratio $E^{j}_{i}$.
In the search stage, we search the number of FFNs $N_{\text{ffn}}$ for every stage $S_i$, and the kernel size $K^{j}_{i}$, as well as the expansion ratios $E^{j}_{i}$ for every block $j$ in stage $i$.

\paragraph{Searching for multi-head self-attention.}
We design our MHSA with weight entanglement \cite{chen2021autoformer}.
We allow the $V$ matrices in each MHSA to have larger dimensions \cite{graham2021levit, tang2023elasticvit}.
More specifically, we search the expansion ratios $E^{j}_{i}$ for the value matrix in every MHSA block, such that
$\texttt{dim}({V_i^j}) = N_{\texttt{head}} \times \texttt{dim}(\texttt{Q-K-V}) \times E^{j}_{i}$, where $N_{\texttt{head}}$ and $\texttt{dim}(\texttt{Q-K-V})$ are fixed.
The dimension of the $Q$ and $K$ matrices are fixed to $N_{\texttt{head}} \times \texttt{dim}(\texttt{Q-K-V})$ such that the attention matrices $A=QK^\intercal$ are shared with all subnets. 
During subnet search, the algorithm finds the optimal subnet by searching the number and the position of MHSA $N_{\texttt{mhsa}}$ and deciding their expansion ratios $E^{j}_{i}$ for the $V$ matrices in the last two stages (\emph{i.e.,} $i \in \{3, 4\}$).
%

\paragraph{Dynamic activation layers.}
In Section \ref{sec:preliminary}, the advanced activation layers such as GELU \cite{hendrycks2016gaussian} are not well-supported by edge devices, thereby incurring a latency overhead during inference.
We designed our supernet to support ReLU \cite{agarap2018deep}, one of the basic activation layers that is hardware-friendly.
During training, we randomly switch between two activation layers for FFN and MHSA.
Our search algorithm searches for the best activation combinations to optimize latency and accuracy.
%

\paragraph{Supernet training algorithm.}
We apply the sandwich rule \cite{yu2019universally} to train our supernet which samples the smallest subnet, the biggest (full) subnet, and $M$ randomly sampled subnets ($M$ = 2 in our experiments).
To accommodate dual FFN, we first sample the FFN structure (\emph{i.e.,} fused or unified) and apply it to the largest and smallest subnets.
For the $M$ randomly sampled subnets, we randomize the FFN selection in every block.
The smallest subnet has minimal width, depth, and kernel size in FFNs, with no MHSAs in the last two layers, using only MHSA downsampling in the third embedding layer.
Conversely, the largest subnet maximizes width, depth, and kernel size, applying MHSAs before all FFNs in layers $S3$ and $S4$.
The FNNs and MHSAs are dropped by drop path \cite{huang2016deep} with probabilities.
We scale the FFN dropping probabilities according to the stage depth so that the last FFN in the stage has the highest probability of being dropped.
All MHSAs are dropped with a constant probability so that the number as well as the position of MHSAs in the stage can be searched.
The full supernet training algorithm is shown in Suppl. \ref{supernet_training_algorithm}.
\section{Searching Subnets for Edge Devices}\label{sec:searching_subnets}

\subsection{Search Objective}
Given a supernet architecture $\mathbf{\mathcal{A}} = \{\alpha_1, ..., \alpha_n \}$ and its trained weight $\mathbf{\mathcal{W}} = \{w_1, ..., w_n \}$, we denote the supernet as $f(\mathbf{\mathcal{A}}, \mathbf{\mathcal{W}})$ and a sampled subnet as $f(\alpha_i, w_i)$, where the architecture $\alpha_i \in \mathbf{\mathcal{A}}$ and the weights $w_i \in \mathbf{\mathcal{W}} $ are sampled from the supernet.
Our search objective is to find an optimal architecture $\alpha^*$ and $w^*$ that maximizes the accuracy while satisfying a set of constraints $\mathcal{C} = \{c_1, ..., c_n\}$ on a given device, such that
\begin{align}
\alpha^*, w^* &= \mathop{\arg\max}_{\alpha \in \mathbf{\mathcal{A}}, w \in \mathbf{\mathcal{W}}} \quad  \text{Acc}(f(\alpha, w)) \nonumber \\
\textrm{s.t.} \quad &  \zeta_i(\alpha, w) < c_i, \quad i=1, 2, ..., n . \label{eq:nas_opt}
\end{align}

 where $\zeta_i$ is the predictor function for the latency or the memory footprint of the subnet.

\subsection{Accuracy Predictor}
Since evaluating thousands of subnets on the validation set during the search process is not practical, we train a neural network $\chi$ to predict the accuracy.
Every block is encoded with a 24-bit length binary string where stages (4-bit), in/output width (8-bit), expansion ratios (6-bit), FFN types (2-bit), kernel sizes (2-bit), and activation functions (2-bit) are one-hot encoded.
The binary string is stacked as a matrix $n \times 24$ and padded to 44 rows for a $n$-block subnet resulting in a $44 \times 24$ matrix.
The network $\chi$ is built with a sequence of 1-D convolution followed by linear layers and it outputs a scalar accuracy prediction.
We train the accuracy predictor using L1 loss, such that
\begin{align*}
\mathop{\arg\min}_{\chi} \quad & |\chi(\alpha_i, w_i) - \text{Acc}(\alpha_i, w_i)| \quad .
\end{align*}
We collect 6k subnet-accuracy pairs on ImageNet \cite{deng2009imagenet} validation set, where the accuracy is evaluated with the weights directly inherited from the supernet (without fine-tuning).
The collected pairs are divided into 5k for training and 1k for validation.
Our goal of the accuracy predictor is to preserve the accuracy rank of the subnets.
%
%
Therefore, we use the best predictor that has the highest Spearman’s rank correlation \cite{lee2021help, dudziak2020brp}.
Our accuracy predictor is simple yet effective and highly preserves the ranking order, as shown in Figure \ref{fig:acc_predictor} (\emph{cf.} Suppl. \ref{quality_of_the_acc_predcitor}).
We use the trained accuracy predictor $\chi(\alpha, w)$ as the proxy that replaces the $\text{Acc}(\alpha, w)$ in Equation \ref{eq:nas_opt}.

\begin{figure}[t]
\centering
\includegraphics[width=0.9\columnwidth]{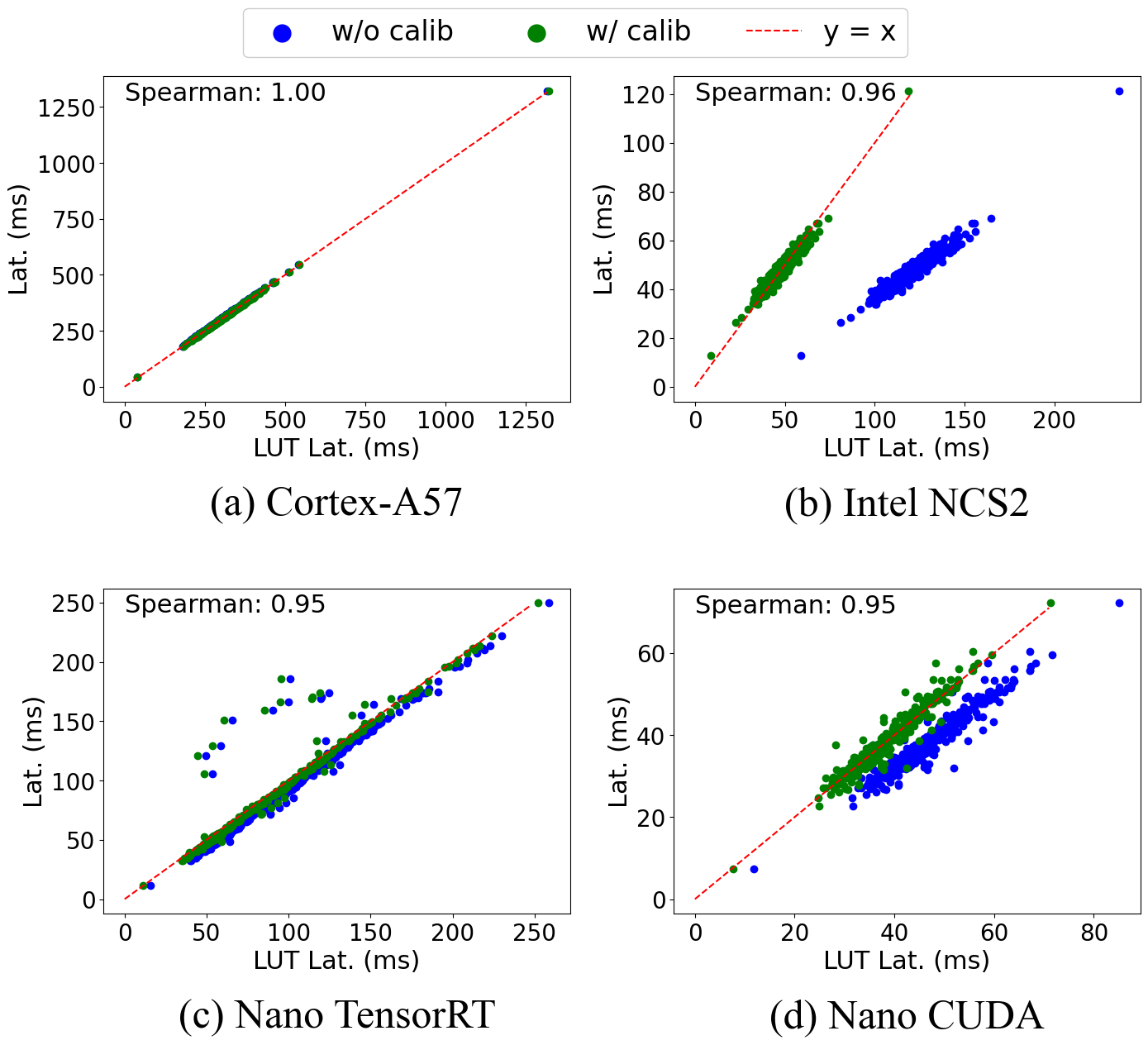}
    \caption{
    Naive latency estimations from block-wised latency lookup tables (LUTs) tend to be overestimated (blue).
    We additionally profile 10 end-to-end subnet latencies to calibrate the LUTs by linear regression.
    We show the high quality of the calibrated latency estimations that fit $y=x$ closely (green).
    }%
    \label{fig:latency_lut}%
\end{figure}

\subsection{Latency Lookup Table with Calibration}
The latency predictor is device-dependent.
Therefore, we build latency lookup tables (LUTs) for every device and profile all possible blocks in every stage on the device.
The total number of blocks is 568 with 4 additional output linear layers for different width choices.
However, simply estimating the end-to-end latency by summing up the latencies of all blocks in the LUT will overestimate the actual latency, since the intermediate feature maps are usually cached in the memory instead of loaded from scratch.
Therefore, for each device, we additionally profile the end-to-end latency of 10 subnets to calibrate the estimation with linear regression,  as shown in Figure \ref{fig:latency_lut}. 
During the search process, we estimate the subnet latency by summing the latency of every block in the lookup table, such that
\begin{equation*}
\Tilde{c_{\text{lat.}}} = \zeta_{\text{lat.}} (\alpha_i, w_i) =\kappa \sum\limits_{j=1}^{n}\text{LUT}(\alpha_i^j) + \epsilon
\end{equation*}
where $\kappa$ and $\epsilon$ are the parameters obtained from the linear regression algorithm.
The $\Tilde{c_{\text{lat.}}}$ is the latency estimation for the $n$-block subnet $\alpha_i$ from the LUT, which should satisfy the latency constraint $c_{\text{lat.}}$ during the search such that $\Tilde{c_{\text{lat.}}} < c_{\text{lat.}}$.
As shown in Figure \ref{fig:latency_lut}, the calibrated block-wise LUTs not only preserve the latency order with a high Spearman’s rank correlation but also have accurate estimations with low error for all devices.
We compare our method with zero-cost proxies in Suppl. \ref{latency_estimation_comparison}.

\begin{figure}[t]
    \centering
    \subfloat[\centering Sampling time]{{\includegraphics[width=4cm]{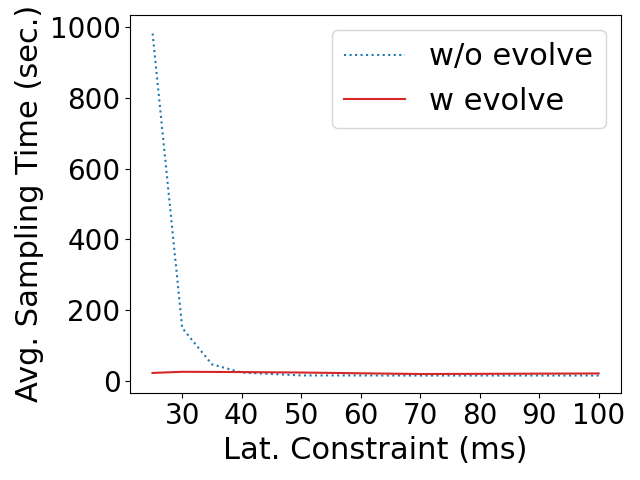} }}%
    \subfloat[\centering Search space evolution]{{\includegraphics[width=4cm]{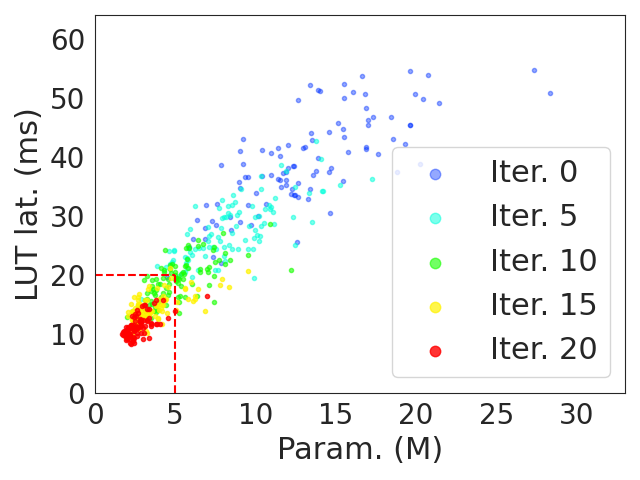} }}%
    \caption{
    (a) The subnet sampling time during the search will increase exponentially if we reduce the latency constraint from $100$ ms to $30$ ms. The sampling is performed on Nano with TensorRT.
    (b) Illustration of search space evolution. Dots represent sampled subnets from the space. After a few interactions, the search space evolves from blue to red dots to where it meets the constraints: $5$ M parameters and $20$ ms latency, \emph{i.e., } red dots are mostly inside the red dashed rectangle.
    }%
    \label{fig:search_space_evolution}%
\end{figure}

\subsection{Hardware-aware Search Space Evolution}
\label{hardware_aware_search_space_evolution}
\paragraph{Search space quality.}
We adopt an evolutionary search algorithm \cite{real2019regularized} to find the optimal subnet for each device.
However, sampling subnets that satisfy the constraints (\emph{e.g.,} latency and memory) from a \emph{very large} search space is difficult (\emph{cf.} Suppl. \ref{search_space_anaylysis}).
The sampling time increases exponentially when we reduce the latency constraint, as shown by the blue line in Figure \ref{fig:search_space_evolution} (a).
%
To address the issue, we update the search space during the search.
The search space evolves if the quality of the search space $\mathbf{\mathcal{Q}}(\mathbf{\mathcal{A}}_t^*)$ defined by the current top $k$ subnets $ \mathbf{\mathcal{A}}^{*}_{t} = \{\alpha_1, ..., \alpha_k\}$ is better than the previous one, such that
\begin{align*}
\mathbf{\mathcal{Q}}(\mathbf{\mathcal{A}}_t^*) - \mathbf{\mathcal{Q}}(\mathbf{\mathcal{A}}_{t-1}^*) > \delta \quad .
\end{align*}

The quality of the search space is evaluated by the average predicted accuracy of the top $k$ subnets, \emph{i.e.}  
\begin{align*}
\mathbf{\mathcal{Q}}(\mathbf{\mathcal{A}}_t^*) = E_{\alpha \in \mathbf{\mathcal{A}}_t^*}[\chi(\alpha)] \quad .
\end{align*}

\begin{figure}[t]
\centering
\includegraphics[width=0.9\columnwidth]{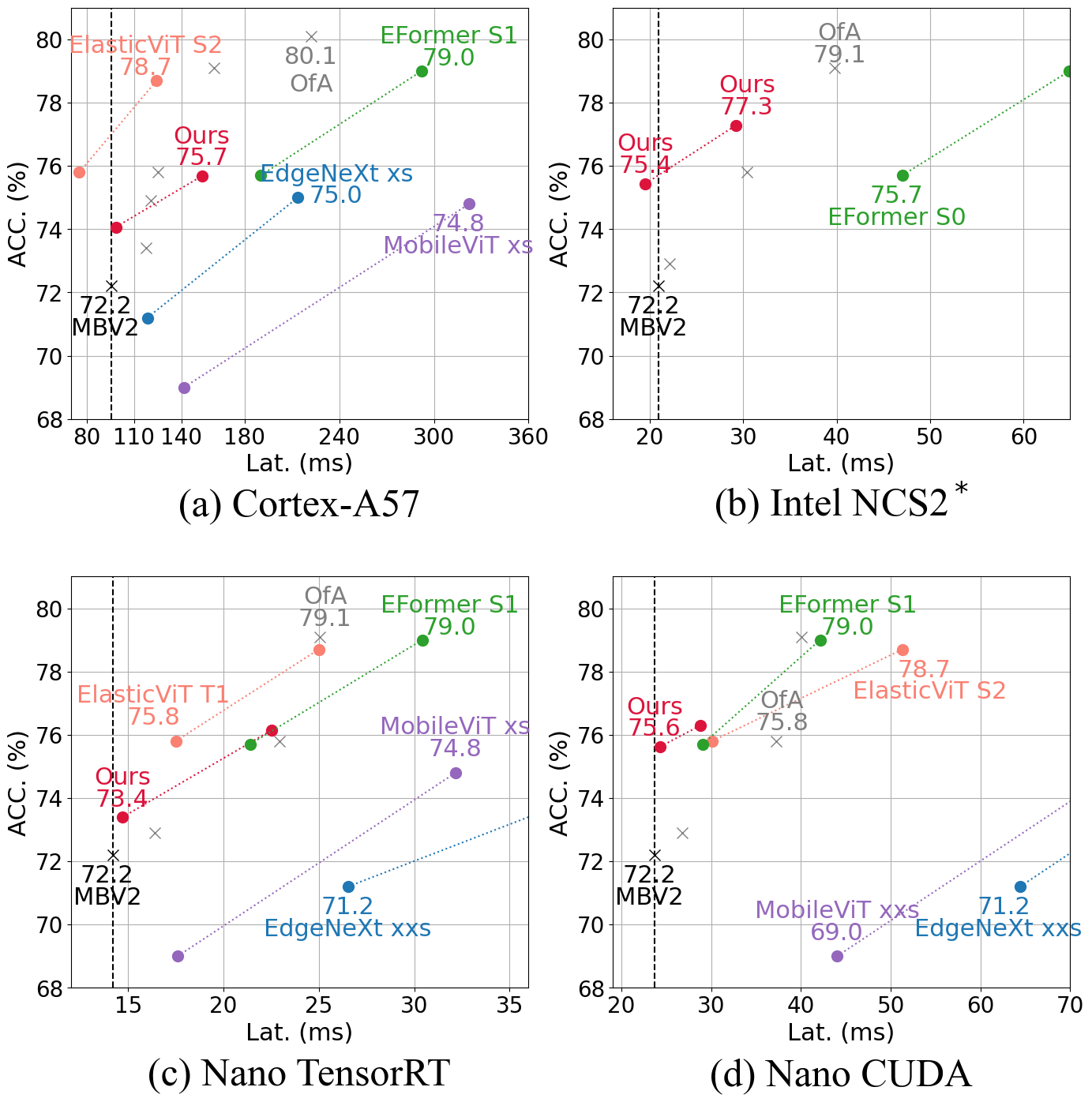}
\caption{Our models achieve MobileNet speed among all hybrid models counterparts across platforms while outperforming MobileNet in accuracy. We also pivot Once-for-all (OfA, Conv only) in grey crosses on the figure for reference.}
\label{fig:results}
\end{figure}

\begin{theorem}
\label{th:mean_probability}
Given $\lambda \in [0, 1]$, the weighted mean of two probability distributions $\Psi $ and $\Phi$ that are defined in the same sample space $\Omega$ such that $\Theta = \lambda\Psi  + (1-\lambda)\Phi$ is a probability distribution defined in $\Omega$. 
\end{theorem}
We prove Theorem \ref{th:mean_probability} in Suppl. \ref{proof_of_the_theorem}.

\paragraph{Search space evolution.}
Based on Theorem \ref{th:mean_probability}, we update the sampling space by the moving average of the probability:
\begin{align*}
p_{t+1}(X^{j}_{i} = x) = \lambda p_{t}(X^{j}_{i} = x) + (1-\lambda)p^{*}(X^{j}_{i} = x) \\
\end{align*}
where $\lambda \in [0, 1]$ is a scalar, $X^{j}_{i}$ is a random variable for a set of architecture choices of block $j$ in stage $i$, $x$ is a specific architecture, and $p_{t}(X^{j}_{i} = x)$ is the current probability of selecting architecture $x$ for the block.
The $p^*$ is a probability defined as
\begin{align*}
p^{*}(X^{j}_{i} = x) = \frac{\sum\limits_{s=1}^{k}\sigma_s(X^{j}_{i} = x)}{k}
\end{align*}
where $\sigma_s$ is an indicator function for subnet $s$ in top $k$ population such that
\begin{align*}
    \sigma_s = 
\begin{cases}
    1,& \text{if } X^{j}_{i} = x\\
    0,              & \text{otherwise}
\end{cases} \quad .
\end{align*}

We maintain separate probability distributions for all stages and update the distribution of each stage at the end of each search episode.
As shown in Figure \ref{fig:search_space_evolution} (b), our search space evolves based on the given constraints and remains constant time subnet sampling.
The detailed algorithm is shown in the Suppl. \ref{searching_algorithm}.
\section{Main Results}\label{sec:main_results}

\begin{table*}[t]
\centering
\small
\caption{The table shows the details of the model comparison.  Our hybrid models outperform MobileNetV2 in accuracy while maintaining the same level of inference latency. Hybrid model counterparts, while having lower FLOPS or model size, fail to achieve MobileNetV2 latency. The naming follows ours\_\{\emph{platform}\}\_\{\emph{compiler}\}@\{\emph{lat}\}ms. We also pivot Once-for-all (OfA, Conv only) for reference.}
\label{tab:result_table}
\begin{tabular}{@{}l|c|ccc|ccc@{}}
\toprule
Method                & Type   & Acc. (\%) & FLOPS (M) & Param. (M) & Cortex (ms) & Nano TRT / Cuda (ms) & NCS2 (ms)                    \\ \midrule \midrule
MobileNetV2           & Conv   & 72.2      & 307.5     & 3.5        & 95.5        & 14.2 / 23.7     & 20.9                            \\ \midrule
ofa-pixel1-40         & Conv   & 74.9      & 259.0     & 6.0        & 120.6       & - \rule{2pt}{0ex} / \rule{2pt}{0ex} -     & -                               \\
ofa-pixel2-35         & Conv   & 73.4      & 224.5     & 5.1        & 117.5       & - \rule{2pt}{0ex} / \rule{2pt}{0ex} -     & -                               \\
ofa-tx2-47            & Conv   & 72.9      & 409.5     & 4.9        & -           & 16.4 / 26.8    & 22.1                            \\ 
ofa-tx2-96            & Conv   & 75.8      & 546.7     & 6.2        & -           & 23.0 / 37.3    & 30.4                            \\ \midrule
MobileViT-XXS\footref{note2}         & Hybrid & 69       & 414.3    & \textbf{1.3}       & 141.7      & 17.6 / 44.0     & -\footref{note1} \\
EdgeNeXt-XXS\footref{note2}          & Hybrid & 71.2     & \underline{259.3}    & \textbf{1.3}       & 118.5      & 26.5 / 64.4    & -\footref{note1}  \\
EdgeViT-XXS           & Hybrid & 74.4      & 555.2    & 4.1         & 162.7      & 31.4 / 67.4   & -\footref{note1}                 \\
ElasticViT T1@224\tablefootnote[3]{We evaluate the accuracy and latency with $224\times224$ inputs. The original implementation is $128\times128$.} & Hybrid &   75.8    &  \textbf{205.1}   &    8.9     &   \textbf{75.4}    &   \underline{17.5} / 30.1     &   -\footref{note1}                              \\
ElasticViT S2     & Hybrid & \textbf{78.7}      & 318.5    & 11.0        & 124.4      & 25.0  / 51.3    & -\footref{note1}                                  \\
EfficentformerV2 S0   & Hybrid & 75.7      & 402.9    & \underline{3.6}         & 190.3      & 21.4  / \underline{29.0}    & 47.0                           \\ \midrule
ours\_cortex@95ms     & Hybrid & 74       & 441.3     & 4.6       & \underline{98.6}        & - \rule{2pt}{0ex} / \rule{2pt}{0ex} -      & -               \\
ours\_cortex@150ms    & Hybrid & 75.7     & 790.8     & 9.7       & 153.0       & - \rule{2pt}{0ex} / \rule{2pt}{0ex} -         & -                               \\
ours\_nano\_trt@13ms  & Hybrid & 73.4     & 806.4     & 7.9       & -           & \textbf{14.7} / \rule{2pt}{0ex} -  \rule{5pt}{0ex}    & -             \\
ours\_nano\_trt@20ms  & Hybrid & 76.1     & 1437.4    & 13.5      & -           & 22.5 / \rule{2pt}{0ex} -  \rule{5pt}{0ex}   & -                                \\
ours\_nano\_cuda@25ms  & Hybrid & 75.6    & 963.0    & 7.0       & -           &  \rule{5pt}{0ex} - \rule{2pt}{0ex} / \textbf{24.3}    & -             \\
ours\_nano\_cuda@30ms  & Hybrid & 76.3    & 1061.6    & 8.8      & -           &  \rule{5pt}{0ex} - \rule{2pt}{0ex}  / 28.8   & -                                \\
ours\_ncs2\_ov@20ms   & Hybrid & 75.4     & 1814.9    & 19.5      & -           & - \rule{2pt}{0ex} / \rule{2pt}{0ex} -    & \textbf{19.2}          \\
ours\_ncs2\_ov@30ms   & Hybrid & \underline{77.3}     & 2921.2    & 36.8      & -           &  - \rule{2pt}{0ex} / \rule{2pt}{0ex} -   & 29.2                            \\ \bottomrule
\end{tabular}
\end{table*}

\subsection{Experimental Setup}
We follow the setup of \citet{li2023rethinking}.
Our models are implemented with PyTorch \cite{paszke2019pytorch} framework and Timm library \cite{timm}.
We use 24 A5000 GPUs to train the supernet for 300 epochs with a total batch size of $3072$ and use 8 A5000 GPUs to fine-tune the searched subnets for 150 epochs with a total batch size of $2048$ on ImageNet 1k training set \cite{deng2009imagenet}.
Models are validated on the ImageNet validation set.
Both training and validation are using the standard resolution $224 \times 224$.
We also use AdamW optimizer \cite{loshchilov2017decoupled}, set the initial learning rate to $10^{-3} \times \text{batch size}/1024$ to train the supernet, and use  $10^{-4} \times \text{batch size}/1024$ to fine-tune the subnets.
The cosine decay is applied in both training and fine-tuning.
RegNetY-16GF \cite{radosavovic2020designing} with 82.9\% top-1 accuracy are used in supernet training and subnet fine-tuning as the teacher model for hard distillation, as \citet{li2023rethinking} and \citet{touvron2021training}.

\paragraph{Commodity devices.} We profile model latency on three different commodity hardware with their compilers using official benchmark tools. \emph{All} models are compiled and profiled with batch size 1.
\begin{itemize}
    \item \textbf{Edge CPU}. We get the model latency on ARM Quad-core Cortex-A57 (Cortex). Models are converted to ONNX \cite{onnx} format and run with the default compiler and execution provider in full precision (FP32).
    \item \textbf{Edge GPU}. We obtain the latency on Jetson Jetson Nano 4G (Nano). Models are converted to ONNX format and compiled by the Cuda / TensorRT (TRT) in full precision (FP32).
    \item \textbf{USB accelerator}. We get the latency on Intel Neural Compute Stick 2 (NCS2). Models are converted to OpenVINO IR (Intermediate Representation) and run with OpenVINO \cite{openvino} in half precision (FP16).
\end{itemize}

\subsection{Image Classification with MobileNet Speed}
The experiments include three different edge devices (Cortex CPU, Nano GPU, and NCS2), four compilers (Default ONNX Execution Provider, Nvidia Cuda and TensorRT, OpenVINO), and two precisions (FP32 and FP16).
In this experiment, we optimize the latency with $224 \times 224$ input resolution that is widely used not only in image classification but also in object detection and segmentation.
We perform the search for each platform and compiler with our trained supernet, accuracy predictor, and pre-built latency tables.
The details of searched architectures are listed in Appendix \ref{searched_architecture_details}.
The weights of the searched subnets are inherited from the supernet and then fine-tuned on the ImageNet with 150 additional epochs.
We test the subnet and report accuracy on the validation set.
To get the latency, models are compiled with the compilers (\emph{e.g.,} TensorRT and OpenVINO) and profiled with a batch size of 1 with $224 \times 224$ resolution input, except for MobileViTs \cite{mehta2021mobilevit} and EdgeNeXt \cite{maaz2022edgenext} which are tested with $256 \times 256$ resolution according to their original implementation \footnote[2]{\label{note2} MobileViTs \cite{mehta2021mobilevit}, EdgeNeXt \cite{maaz2022edgenext} are trained and tested with $256 \times 256$ according to their original implementation}.

We show the results in Figure \ref{fig:results} and list the details of the comparison in Table \ref{tab:result_table}.
%
%
Once-for-all (OfA, Conv only) is also shown in Table \ref{tab:result_table} and Figure \ref{fig:results} for reference only.
The models searched from our framework are named ours\_\{\emph{platform}\}\_\{\emph{compiler}\}@\{\emph{lat}\}ms where \emph{lat} is the search latency constraint.
For Cortex-A57, we use only the ONNX default compiler, so we omit \{\emph{compiler}\} in Table \ref{tab:result_table}.
Prior-art \cite{mehta2021mobilevit, pan2022edgevits, maaz2022edgenext, li2023rethinking} \emph{fails to reach MobileNetV2 latency with on-par accuracy} on the three hardware platforms\footnote[1]{\label{note1} MobileViTs \cite{mehta2021mobilevit}, EdgeNeXt \cite{maaz2022edgenext}, EdgeViT \cite{pan2022edgevits}, ElasticViT \cite{tang2023elasticvit} fail to compile on Intel NCS2 due to unsupported operators.}, although they have fewer FLOPS (EdgeNeXt) or smaller model size (MobileViTs, EdgeNeXt) than MobileNetV2.
Although ElasticViT T1\cite{tang2023elasticvit} has the lowest latency on Cortex-A57, it uses a search space for edge CPU (\emph{e.g.,} MobilenetV2 and V3 block) and thereby fails to reach MobilenetV2 on other computation platforms.
Our hybrid models outperform MobileNetV2 in accuracy (\textbf{74\%} \emph{vs.} \textbf{72.2\%} on Cortex-A57) while maintaining the same level of inference latency (\textbf{98.6 ms} \emph{vs.} \textbf{95.5 ms} on Cortex-A57).
Our study shows that zero-cost proxies (\emph{e.g.,} model size and FLOPS) \emph{fail to} reflect the actual on-device latency.
From the experiments, the key to optimizing model architecture for devices is incorporating device-friendly operators in the search space (\emph{e.g.,} convolution and activation choices) and a low-cost accurate latency predictor (\emph{e.g.,} calibrated LUTs).
%
%
We expect that our search algorithm can identify more competitive model architectures once the MHSA and GELU are optimized for the target hardware.
%

\subsection{Joint Optimization of Latency and Model Size}
The smaller models benefit from using the limited memory on edge devices and being easily downloaded from cloud to edge devices.
This motivates us to jointly optimize the latency and model size.
We experiment with joint optimization on ARM Cortex-A57 and Nvidia Jetson Nano platforms by constraining the search algorithm with both latency and number of parameters.
In the experiment, we use ONNX default compiler for Cortex-A57 and TensorRT for Jetson Nano.
As shown in Figure \ref{fig:search_space_evolution} (b), our search space evolves to where it meets both constraints.
The searched models follow the naming @\{\emph{lat}\}ms\_@\{\emph{size}\}M, where \emph{lat} and \emph{size} are the search constraints.
As shown in Table \ref{tab:joint_opt}, our framework searches optimal models with the given constraints for Cortex-A57 and Nano with little accuracy loss ($0.2\%$).

\begin{table}[t]
\centering
\scriptsize
\caption{Joint optimization latency and model size for Cortex-A57 with default compiler, Nano with TensorRT compiler, and NCS2 with OpenVINO compiler. The naming follows @\{\emph{lat}\}ms\_@\{\emph{size}\}M}
\label{tab:joint_opt}
\begin{tabular}{@{}l|cc|c@{}}
\toprule
Method               & Acc. (\%) & Param. (M) & cortex / nano / ncs2 \\ \midrule \midrule
MobileNetV2          & 72.2                                                 & 3.5                                                   & 95.5 / 14.2  / 20.9                                               \\
EfficientFormerV2 S0 & 75.7                                                 & 3.6                                                   & 190.3 / 21.4 / 47.0                                               \\ \midrule
\multicolumn{4}{c}{Cortex-A57}                                                                                                                                                                          \\ \midrule
@150ms               & 75.7                                                 & 9.7                                                   & 153.0 / \rule{4pt}{0ex} - \rule{4pt}{0ex} / \rule{4pt}{0ex}- \rule{4pt}{0ex}                                                    \\
@150ms\_param@5M     & 75.5                                                 & 5.0                                                   & \textbf{152.4} / \rule{4pt}{0ex} - \rule{4pt}{0ex} / \rule{4pt}{0ex} - \rule{4pt}{0ex}                                                     \\ \midrule
\multicolumn{4}{c}{Nano TensorRT}                                                                                                                                                                       \\ \midrule
@20ms                & 76.1                                                 & 13.5                                                  & \rule{4pt}{0ex} - \rule{4pt}{0ex}  / 22.5 / \rule{4pt}{0ex} - \rule{4pt}{0ex}                                                      \\
@20ms\_param@5M      & \textbf{75.9}                                                 & 5.0                                                   & \rule{4pt}{0ex} - \rule{4pt}{0ex}  / 22.2 / \rule{4pt}{0ex} - \rule{4pt}{0ex}                                                       \\ \midrule
\multicolumn{4}{c}{NCS2 OpenVINO}                                                                                                                                                                       \\ \midrule
@45ms                & 78.6                                                 & 47.9                                                  & \rule{4pt}{0ex} - \rule{4pt}{0ex}  /  \rule{4pt}{0ex} - \rule{4pt}{0ex}  / 47.5                                                      \\
@45ms\_param@5M      & \textbf{75.7}                                                 & 5.0                                                   & \rule{4pt}{0ex} - \rule{4pt}{0ex}  /  \rule{4pt}{0ex} - \rule{4pt}{0ex}  / \textbf{43.0}                                                      \\ \bottomrule
\end{tabular}
\end{table}

\subsection{Downstream Tasks via Transfer Learning}
We perform transfer learning from the ImageNet pre-trained weight to various downstream tasks: CIFAR10, CIFAR100 \cite{krizhevsky2009learning}, Food \cite{bossard2014food}, and Pets \cite{parkhi2012cats}.
All models are trained for 50 epochs with downstream datasets on an A500 GPU and set the batch size to $256$ with a $10^{-3}$ base learning rate scaled by the batch size.
The results are shown in Table \ref{tab:transfer_learning}.
In general, our models outperform in accuracy their counterparts with similar latency in the downstream classification tasks on Cortex-A57.
Moreover, our models match the MobileNetV2 latency on Cortex-A57 CPU (\emph{cf.} Table \ref{tab:result_table}).

\begin{table}[h!]
\centering
\small
\caption{Our models have higher accuracy in the downstream classification tasks and better (lower) latency on Cortex-A57 than other mobile-friendly models.}
\label{tab:transfer_learning}
\begin{tabular}{@{}l|cccc@{}}
\toprule
Method                              & CF10 & CF100 & Food & Pets   \\ \midrule \midrule
MobileNetV2                         & 91.6 & 72.1 & 71.2 & 83.2   \\
MobileViT XXS\footref{note2}        & 93.1 & 71.2  & 73.1 & 80.7   \\ 
EdgeNeXt XXS\footref{note2}         & 95.7 & 76.2  & 80.0 & 84.2   \\ \midrule
cortex@95ms          & 97.1 & 80.9  &  \textbf{82.7}    &  86.0      \\
cortex@150ms         & \textbf{97.3} & \textbf{81.1}  &  81.9    &  \textbf{86.9}       \\ \bottomrule
\end{tabular}
\end{table}

\section{Object Detection}
We integrate our searched subnets as the backbone to SSDLite \cite{sandler2018mobilenetv2}.
We train the SSDLite with different backbones on COCO2017 dataset \cite{lin2014microsoft} by using the MMDetection library \cite{mmdetection}.
We load the ImageNet pre-trained weights of each backbone, and train the detection models with a resolution of $320 \times 320$.
The learning rates are tuned for each model.
We deploy models on the Nvidia Jetson Nano 4G by using the MMDeploy library \cite{mmdeploy} and profile the latency with Nvidia TensorRT profiling tools. 
As shown in Table \ref{tab:object_detection}, our model outperforms MobileViT XXS and EdgeNeXt XXS in both mAP and latency.

\begin{table}[h!]
\centering
\caption{The SSDLite with our searched model reaches the highest mAP and lowest end-to-end latency on Nano among all hybrid alternatives on COCO2017 object detection dataset.}
\label{tab:object_detection}
\begin{tabular}{@{}ccc@{}}
\toprule
SSDLite Backbone      & mAP  & Lat. (ms)   \\ \midrule \midrule
MobileNetV2   & 20.5 & 54.4  \\
MobileViT XXS & 19.1 & 65.4  \\
EdgeNeXt XXS  & 19.3 & 73.3 \\ \midrule
Nano\_trt@13ms (Ours) & 22   & 60.6  \\ \bottomrule
\end{tabular}
\end{table}

\section{Conclusion}
\label{conclusion}
We propose a unified NAS framework that searches for hybrid networks with MobileNetV2-speed yet superior accuracy for low-cost commodity edge devices.
Our framework incorporates different device-friendly operations for diverse edge devices with different hardware designs.
The proposed search algorithm relies on real latency instead of zero-cost proxies (\emph{e.g.,} FLOPs, number of parameters) and reduces the sampling time by search space evolution for a wide variety of edge devices.
Our experiments show our hybrid models match MobileNetV2-speed on Edge CPUs, Edge GPUs, and USB accelerators with better accuracy than MobileNetV2.
%

\clearpage
\balance
\bibliography{egbib}

\clearpage
\balance

\setcounter{figure}{0}
\setcounter{table}{0}
\setcounter{section}{0}
\renewcommand{\thefigure}{S\arabic{figure}}
\renewcommand{\thetable}{S\arabic{table}}
\renewcommand{\thesection}{S\arabic{section}}

\twocolumn[
\begin{center}
\textbf{\Large Supplemental Materials}
\end{center}
]

\section{Proof of Theorem 1}
\label{proof_of_the_theorem}
\begin{theorem}
Given $\lambda \in [0, 1]$, the weighted mean of two probability distributions $\Psi $ and $\Phi$ that are defined in the same sample space $\Omega$ such that $\Theta = \lambda\Psi  + (1-\lambda)\Phi$ is also a probability distribution defined in $\Omega$. 
\end{theorem}

\begin{proof}
Since $\Psi $ and $\Phi$ are defined in the same sample space $\Omega$, we define their associated density distribution functions are $p_{\Psi }$ and $p_{\Phi}$ such that.
\begin{align*}
    p_{\Psi }(X=x) &= q_{\Psi}^x, \quad \text{and} \quad \int_{\Omega} p_{\Psi }(X=x) = 1 \\
    p_{\Phi }(X=x) &= q_{\Phi}^x,  \quad \text{and} \quad \int_{\Omega} p_{\Phi}(X=x) = 1  \quad .
\end{align*}
The $X$ is the random variable of the sample space $\Omega$, and $q^x$ is the probability that associated with $X=x, x\in \Omega$.

Given $\lambda \in [0, 1]$, we have
\begin{align*}
    p_{\Theta }(X=x) &= \lambda p_{\Psi }(X=x) + (1-\lambda) p_{\Phi }(X=x) \\
    \int_{\Omega} p_{\Theta }(X=x) &= \int_{\Omega} \lambda p_{\Psi }(X=x) + \int_{\Omega} (1-\lambda) p_{\Phi }(X=x) \\ 
    \int_{\Omega} p_{\Theta }(X=x) &= \lambda + (1-\lambda) = 1 \quad . 
\end{align*}
This concludes the proof that $\Theta$ is also a probability distribution defined in the sample space $\Omega$.
\end{proof}

\begin{algorithm} [h!]
\caption{The supernet training algorithm}\label{alg:supernet_training}
\begin{algorithmic}
\Require $total\_epoch$, \text{supernet} $\mathbf{\mathcal{A}}$,  \text{training set} $\mathbf{\mathcal{D}}$
\For{$epoch < total\_epoch$}
\For{$ ( \textbf{X}, \textbf{Y} ) \in \mathbf{\mathcal{D}}$}
\State \(\triangleright\) Sample an FFN type for the max and min
\State $FFN(\text{max}) = \texttt{Sample}(\{\text{Unified}, \text{Fused}\}) $
\State $FFN(\text{min}) = \texttt{Sample}(\{\text{Unified}, \text{Fused}\}) $
\State
\State \(\triangleright\) Sample random subnet from $\mathbf{\mathcal{A}}$
\State $\text{rand-1}, \text{rand-2} \in \texttt{Sample}(\mathbf{\mathcal{A}})$
\State
\State \(\triangleright\) Sandwich rule
\For{$ f \in \{\text{max}, \text{rand-1}, \text{rand-2}, \text{min}\}$}
\State $\textbf{L} \gets \texttt{Criterion} (\textbf{Y}, f(\textbf{X}))$
\State $\texttt{Backprop}(\textbf{L})$ \Comment{accumulate gradients at leaf}
\EndFor
\State $\mathbf{\mathcal{A}} = Optimizer.\texttt{step}(\mathbf{\mathcal{A}})$ \Comment{Update parameters}
\EndFor
\EndFor
\end{algorithmic}
\label{supernet_training_algorithm}
\end{algorithm}

\section{Dual \emph{vs.} Entangled FFNs}
\label{dual_vs_entangled_ffn}
\begin{figure}[h]
\centering
\includegraphics[width=.9\columnwidth]{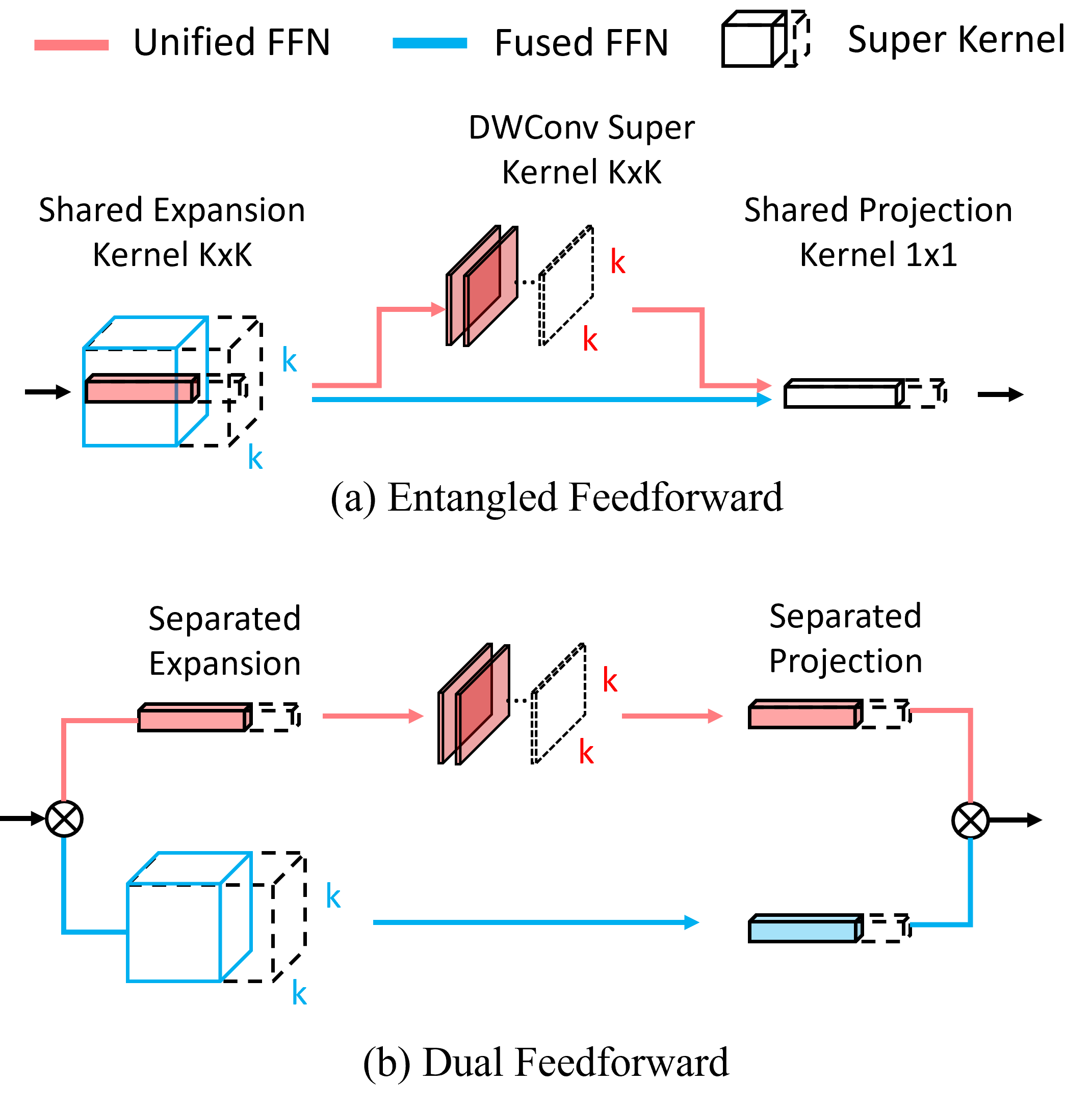}
\caption{The supernet design of (a) Entangled FFN  and (b) Dual FFN.}
\label{fig:entangled_vs_dual}
\end{figure}

\begin{figure}[h]
    \centering
    \subfloat[\centering Training loss]{{\includegraphics[width=4cm]{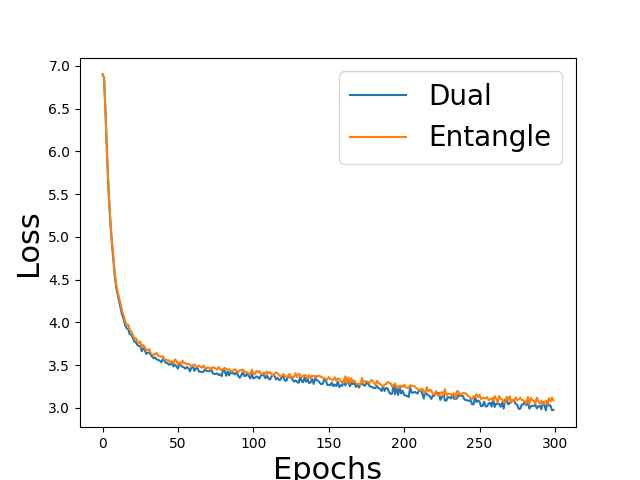} }}%
    \subfloat[\centering Validation accuracy]{{\includegraphics[width=4cm]{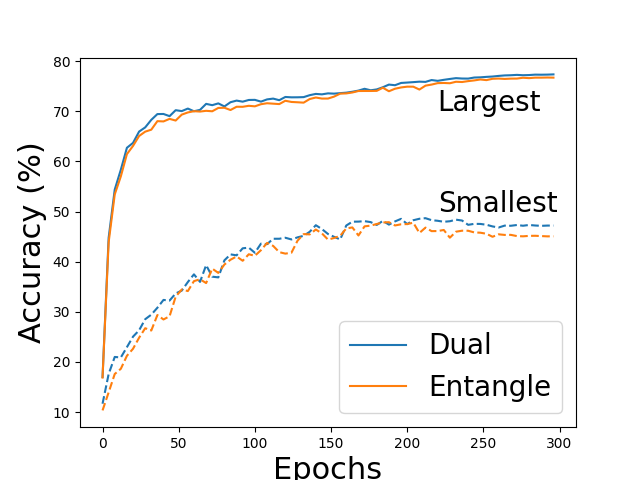} }}%
    \caption{
     Dual \emph{vs.} Entangled FFNs.
    }%
    \label{fig:entangled_vs_dual_loss_valid}%
\end{figure}

We show two different designs of the supernet to accommodate two types of FFN in Figure \ref{fig:entangled_vs_dual}.
The dual feedforward networks are composed of two sets of separated weight matrices, while entangled feedforward networks share the expansion and projection weight matrices.
Only one type of FFN will be activated (unified or fused) in a block during the training time.
We empirically find the dual FFN converges slightly better than entangled FFN and has higher accuracy of the subnets, as shown in Figure \ref{fig:entangled_vs_dual_loss_valid}.
We use dual FFN in the supernet for all experiments.

\section{Supernet Training Algorithm}
We show the pseudo-code of our supernet training algorithm in Algorithm \ref{supernet_training_algorithm}.

\begin{figure*}[t]
\centering
\includegraphics[width=2\columnwidth]{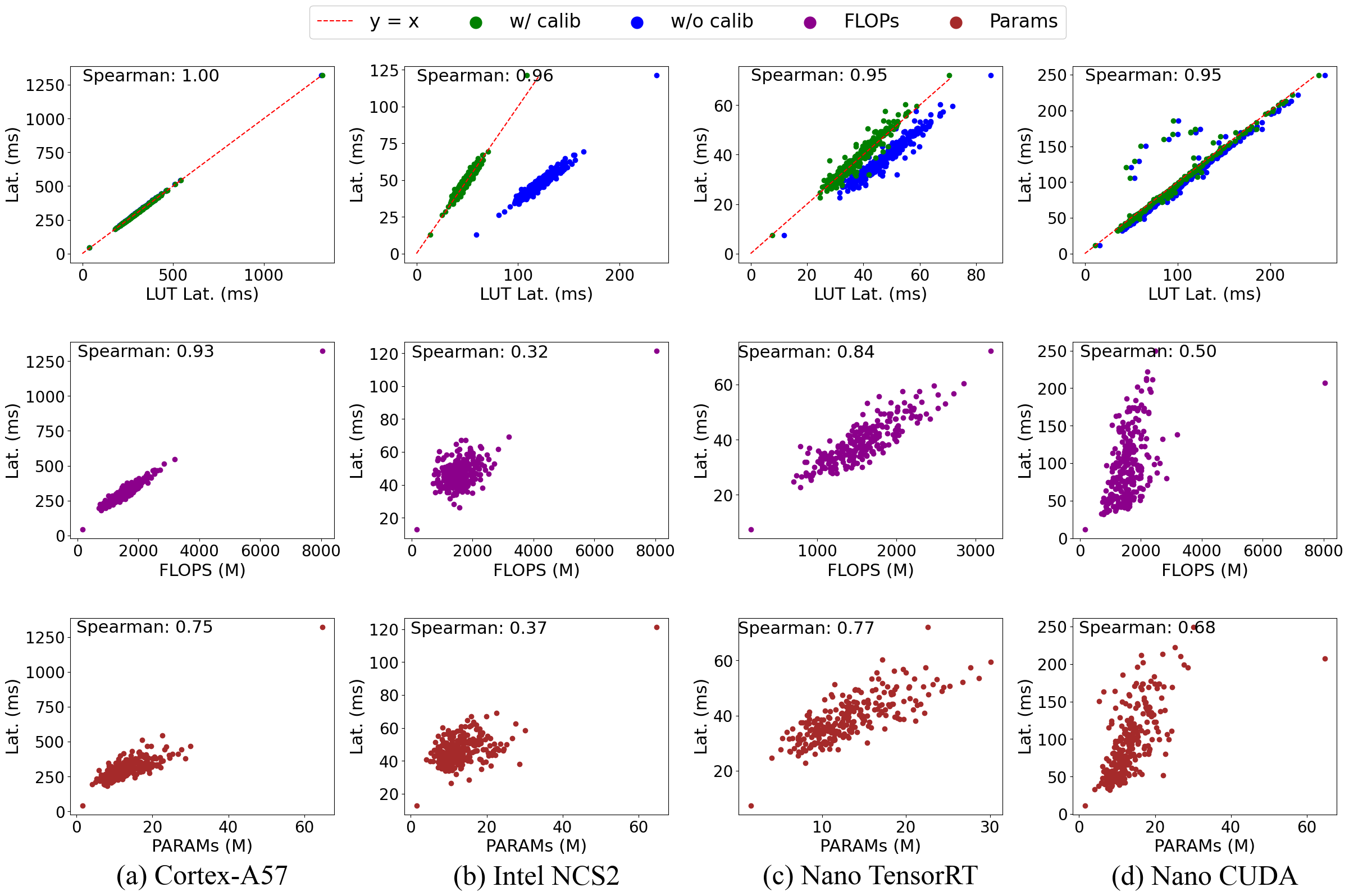}
\caption{We compare the latency estimations with lookup tables (LUTs; top), number of floating-point operations (FLOPS; middle), and number of parameters (Params; bottom).
We calibrate the LUTs with an additional 10 subnet end-to-end latency (Green points). 
Best viewed in color.
}
\label{fig:latency_estimation_comparison}
\end{figure*}

\section{Latency Estimation Comparison}
\label{latency_estimation_comparison}
We compare our method with zero-cost proxies in Figure \ref{fig:latency_estimation_comparison}.
The results show that the zero-cost proxies such as the number of floating-point operations (FLOPS) and the number of parameters do not reflect the real latency on diverse devices.
We profile all possible blocks on the devices and build latency lookup tables (LUTs).
However, LUTs overestimate the latency estimation since the feature maps are usually cached on the device.
To this end, we additionally profile 10 random subnets to calibrate our latency LUTs.
The resulting latency estimations from the calibrated lookup tables not only highly preserve the latency order (high Spearman’s rank correlation), but also accurately estimate the subnet on-device latency (The green dots fit $y=x$ line closely).

We compare our latency predictor with the state-of-the-art machine-learning-based predictor, nn-Meter \cite{zhang2021nn}.
We random sample 100 convolutional neural networks and 100 hybrid networks.
We profile the ground-truth latency on the Nvidia Jetson Nano 4G and estimate the latency with our lookup tables and nn-Meter.
The RMSE of the latency in milliseconds is presented in Table \ref{tab:latency_predictor_comparison}
Note that nn-Meter is only applicable to convolutional neural networks.
nn-Meter utilizes machine learning algorithms such as random forests to predict the latency for each computation node in the architecture graph, which can be expensive.
Our method looks up the latency for each layer and scales the summed latency by the scaling factors.
Thus, our method incurs nearly zero computational overhead during the search.
We report the average of 100 prediction time on an AMD EPYC 7413 24-Core Processor, a server-class CPU, in Table \ref{tab:latency_predictor_comparison}.

\begin{table}[h]
\centering
\caption{We compare the RMSE of the predicted latency from LUTs and nn-Meter against the ground-truth latency. Additionally, we report the runtime of both methods on a server-class CPU. }
\label{tab:latency_predictor_comparison}
\begin{tabular}{@{}c|cc|c@{}}
\toprule
Method   & Conv Only & Hybrid & Runtime (ms) \\ \midrule
LUT      & 3.99      & 2.09   & 0.25            \\
nn-Meter & 2.44      & --     & 378.97          \\ \bottomrule
\end{tabular}
\end{table}

\section{The Quality of the Accuracy Predictor}
\label{quality_of_the_acc_predcitor}
\begin{figure}[h!]
\centering
\includegraphics[width=0.9\columnwidth]{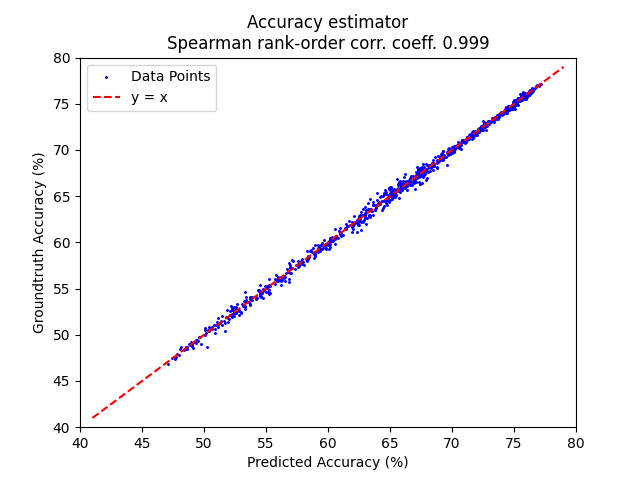}
\caption{The quality of the accuracy predictor}
\label{fig:acc_predictor}
\end{figure}

The accuracy predictor is trained with $6000$ subnet-accuracy pairs, where the accuracy is evaluated with the weights directly inherited from the supernet on the ImageNet validation set.
We show the correlations between ground truth accuracy ($y$-axis) and predicted accuracy ($x$-axis) in Figure \ref{fig:acc_predictor}.
Our accuracy predictor preserves the accuracy rank of the subnets.
We use the value predicted from our accuracy predictor as a proxy instead of exhaustively evaluating subnets on the dataset during the search.

\section{Analysis of Subnet Accuracy}
\label{subnet_accuracy}
We show the predicted accuracy of the subnets from our accuracy predictor, the accuracy of the subnets inherited from the supernet, and fine-tuned accuracy of the subnets in Table \ref{tab:accuracy_comparison} for reference.
We note that our accuracy predictor is trained with $6000$ subnet-accuracy pairs, where the accuracies are evaluated with the weights directly inherited from the supernet.

\begin{table}[h]
\centering
\small
\caption{Accuracy comparison of subnets.}
\label{tab:accuracy_comparison}
\begin{tabular}{@{}l|cc|cc@{}}
\toprule
Models         & Pred. (\%) & Inherit. (\%) & FT (\%) \\ \midrule\midrule
cortex@95ms    & 64.4  & 68       & 74.1  \\
cortex@150ms   & 68.2  & 70.3     & 75.7  \\ \midrule
nano\_trt@13ms & 64.3  & 66.1     & 73.4  \\
nano\_trt@20ms & 70.3  & 69.1     & 76.1  \\ \bottomrule
\end{tabular}
\end{table}


\begin{algorithm}[th]
\caption{Evolutionary search. }\label{alg:evolutionary_search}
\begin{algorithmic}
\Require $pop\_size, total\_gen, \text{constraints } \zeta$
\State $prob \gets \{p, ...\}$ \Comment{initial search space probabilities}
\State $pop \gets \{\}$ \Comment{initial empty population}
\State $history \gets \{\}$ \Comment{initial empty history}
\State \(\triangleright\) Random sample subnets
\While{$|pop| < pop\_size$}
\State $subnet \gets \texttt{Sample}(\mathbf{\mathcal{A}}, prob)$  \Comment{Sample a subnet}
\State $history.\texttt{add}(subnet)$ 
\If {$\texttt{check}(subnet, \zeta)$} \Comment{check constraints}
\State $pop.\texttt{add}(subnet)$
\EndIf
\State \(\triangleright\) Search space evolution at initialization stage
\If {$|history| \bmod evolve\_step = 0$}
\State $topk \gets sort(history, \zeta)$ \Comment{sorted by constraints}
\State $prob \gets \lambda \,prob + (1-\lambda )\, \texttt{get\_prob}(topk)$
\EndIf
\EndWhile
\State
\State \(\triangleright\) Start searching for $total\_gen$ generations
\State $\mathbf{\mathcal{Q}} \gets \texttt{space\_quality}(pop)$
\While{$gen < total\_gen$}
\State $topk \gets \texttt{sort}(pop, \zeta)$ \Comment{sorted by constraints}
\State $\mathbf{\mathcal{Q}}_{t} \gets \texttt{space\_quality}(topk)$
\State \(\triangleright\) Update the search space
\If {$\mathbf{\mathcal{Q}}_{t} - \mathbf{\mathcal{Q}} > \delta$}
\State $\mathbf{\mathcal{Q}} \gets \mathbf{\mathcal{Q}}_{t}$
\State $prob \gets \lambda \,prob + (1-\lambda )\, \texttt{get\_prob}(topk)$
\EndIf
\State
\State \(\triangleright\) Update population
\State $pop \gets topk$
\State $pop.\texttt{update}(\texttt{mutate}(topk, prob))$
\State $pop.\texttt{update}(\texttt{crossover}(topk, prob))$
\State \(\triangleright\) Random sample subnets
\While{$|pop| < pop\_size$}
\State $subnet \gets \texttt{Sample}(\mathbf{\mathcal{A}}, prob)$
\If {$\texttt{check}(subnet, \zeta)$}
\State $pop.\texttt{add}(subnet)$
\EndIf
\EndWhile
\State $history.\texttt{update}(pop)$ \Comment{update population history}
\EndWhile
\State $\Return \; \texttt{sort}(pop, \zeta)[0]$ \Comment{return top1 subnet}
\end{algorithmic}
\end{algorithm}

\section{Analysis of Search Space}
\label{search_space_anaylysis}
Our search space contains roughly $10 ^{45}$ subnets, which is much larger than previous work \cite{cai2020once}.
The details of the search space are shown in Table 2 in the paper.
We show the mathematical analysis of our search space in the section.
Our supernest consists of a stem conv followed by 4 stages with 3 embeddings in between.
\begin{itemize}
    \item Each stage has 4 choices of width values resulting in $4^4$ width combinations.
    \item We search the activation layers for stem conv resulting in $2$ architecture choices.
    \item The embedding layers only consist of one layer of convolution and do not contain activation layers, except for the MHSA downsampling layer, which has 6 architecture choices.
    \item In the first and second stages, we have $24$ choices for each FFN block and 2 depth choices (2 or 3 blocks) for each stage and therefore the total number of choices of the architecture is $\sum_{n=2}^{N=3} 24^n$.
    \item In the third and fourth stages, we have $24$ choices for each FFN block for each stage. We search for the number $m$ and the position of MHSA among $n$ FFN blocks each of which has 6 choices resulting in $6^m\binom{n}{m}$.
\end{itemize}
Summarizing the above items together in an equation, we obtain 
\begin{align*}
4^4 \times 2 \times
\sum_{n=2}^{3}24^n \times \sum_{n=2}^{3} 24^n \times \\
\sum_{n=6}^{9} 24^n(\sum_{m=0}^n 6^m\binom{n}{m}) \times 
6 \times \\
\sum_{n=4}^{6} 24^n(\sum_{m=0}^n 6^m\binom{n}{m}) \approx 10 ^{45}
\end{align*}


\section{The search algorithm}
\label{searching_algorithm}
We show the pseudo-code of our search algorithm in Algorithm \ref{alg:evolutionary_search}.
The top-k subnets perform mutation and crossover in the search.
Our search space evolves based on the current top-ranking subnets.
After some generations, the algorithm returns the top1 subnet architecture.

\section{Searched Architecture Details}
\label{searched_architecture_details}
We show the architecture details of searched subnets in Table \ref{tab:subnet_details_cortex} and Table \ref{tab:subnet_details_nano_trt}.
Our search algorithm tends to adopt depthwise convolutions (unified FFNs) with large-size kernels (\emph{e.g.,} 5) and MHSA for ARM Cortex-A57.
In contrast, our search algorithm is in favor of vanilla convolutions (fused FFNs)  with small-size kernels (\emph{e.g.,} 3) over depthwise convolutions (unified FFNs) and MHSA for Nvidia Jetson Nano 4G with TensorRT compiler.
GELU and MHSA which boost accuracy are placed in the model closer to the output with minimal latency impact.
The result shows that our search optimizes the model architecture for different hardware implementations and characteristics.
Therefore, we expect that our search algorithm can propose more competitive model architectures once the MHSA and GELU are optimized for the target hardware.

\begin{table*}[h]
\centering
\small
\caption{
Architecture details of searched subnets on ARM Cortex-A57. We abbreviate output resolution (Res.), output channel width (Ch.), fused FFNs (Fused), Unified FFNs (unified), and Multi-head self-attention layers (MHSA), embedding layers (Embed 3), activation layers (A), GELU (G),  ReLU (R), expansion ratio (E), and kernel size (K) in the Table.
}
\label{tab:subnet_details_cortex}
\begin{tabular}{?cc?ccc?ccc?ccc?}
\toprule
\multicolumn{2}{?l?}{}                                                   & \multicolumn{3}{c?}{Cortex@150ms}                                                    & \multicolumn{3}{c?}{Cortex@150ms\_@5M}                                               & \multicolumn{3}{c?}{Cortex@95ms}                                                     \\ \toprule
\multicolumn{1}{?c|}{Stage}                     & Res.               & \multicolumn{1}{c|}{Ch.}              & \multicolumn{1}{c|}{Type}    & E / K / A & \multicolumn{1}{c|}{Ch.}              & \multicolumn{1}{c|}{Type}    & E / K / A & \multicolumn{1}{c|}{Ch.}              & \multicolumn{1}{c|}{Type}    & E / K / A \\ \hline
\multicolumn{1}{?c|}{Conv}                 & 1/4                    & \multicolumn{1}{c|}{32}                   & \multicolumn{1}{c|}{Conv}    & - / 3 / R & \multicolumn{1}{c|}{32}                   & \multicolumn{1}{c|}{Conv}    & - / 3 / R & \multicolumn{1}{c|}{24}                   & \multicolumn{1}{c|}{Conv}    & - / 3 / R \\ \hline
\multicolumn{1}{?c|}{\multirow{2}{*}{Stage 1}}  & \multirow{2}{*}{1/4}   & \multicolumn{1}{c|}{\multirow{2}{*}{32}}  & \multicolumn{1}{c|}{Fused}   & 2 / 3 / R & \multicolumn{1}{c|}{\multirow{2}{*}{32}}  & \multicolumn{1}{c|}{Fused}   & 3 / 3 / R & \multicolumn{1}{c|}{\multirow{2}{*}{24}}  & \multicolumn{1}{c|}{Fused}   & 2 / 3 / R \\ \cline{4-5} \cline{7-8} \cline{10-11} 
\multicolumn{1}{?c|}{}                          &                        & \multicolumn{1}{c|}{}                     & \multicolumn{1}{c|}{Fused}   & 2 / 3 / R & \multicolumn{1}{c|}{}                     & \multicolumn{1}{c|}{Fused}   & 3 / 3 / R & \multicolumn{1}{c|}{}                     & \multicolumn{1}{c|}{Fused}   & 2 / 3 / R \\ \hline
\multicolumn{1}{?c|}{Embed 1}                   & 1/8                    & \multicolumn{1}{c|}{40}                   & \multicolumn{1}{c|}{Conv}    & -         & \multicolumn{1}{c|}{64}                   & \multicolumn{1}{c|}{Conv}    & -         & \multicolumn{1}{c|}{40}                   & \multicolumn{1}{c|}{Conv}    & -         \\ \hline
\multicolumn{1}{?c|}{\multirow{2}{*}{Stage 2}}  & \multirow{2}{*}{1/8}   & \multicolumn{1}{c|}{\multirow{2}{*}{40}}  & \multicolumn{1}{c|}{Fused}   & 2 / 3 / R & \multicolumn{1}{c|}{\multirow{2}{*}{64}}  & \multicolumn{1}{c|}{Unified} & 2 / 3 / R & \multicolumn{1}{c|}{\multirow{2}{*}{40}}  & \multicolumn{1}{c|}{Unified} & 2 / 3 / R \\ \cline{4-5} \cline{7-8} \cline{10-11} 
\multicolumn{1}{?c|}{}                          &                        & \multicolumn{1}{c|}{}                     & \multicolumn{1}{c|}{Fused}   & 2 / 3 / R & \multicolumn{1}{c|}{}                     & \multicolumn{1}{c|}{Fused}   & 2 / 3 / R & \multicolumn{1}{c|}{}                     & \multicolumn{1}{c|}{Fused}   & 2 / 3 / R \\ \hline
\multicolumn{1}{?c|}{Embed 2}                   & 1/16                   & \multicolumn{1}{c|}{96}                   & \multicolumn{1}{c|}{Conv}    & -         & \multicolumn{1}{c|}{96}                   & \multicolumn{1}{c|}{Conv}    & -         & \multicolumn{1}{c|}{96}                   & \multicolumn{1}{c|}{Conv}    & -         \\ \hline
\multicolumn{1}{?c|}{\multirow{10}{*}{Stage 3}} & \multirow{10}{*}{1/16} & \multicolumn{1}{c|}{\multirow{10}{*}{96}} & \multicolumn{1}{c|}{Unified} & 3 / 3 / R & \multicolumn{1}{c|}{\multirow{10}{*}{96}} & \multicolumn{1}{c|}{Unified} & 3 / 5 / R & \multicolumn{1}{c|}{\multirow{10}{*}{96}}  & \multicolumn{1}{c|}{Unified} & 3 / 3 / R \\ \cline{4-5} \cline{7-8} \cline{10-11} 
\multicolumn{1}{?c|}{}                          &                        & \multicolumn{1}{c|}{}                     & \multicolumn{1}{c|}{MHSA}    & 2 / - / R & \multicolumn{1}{c|}{}                     & \multicolumn{1}{c|}{MHSA}    & 2 / - / R & \multicolumn{1}{c|}{}                     & \multicolumn{1}{c|}{MHSA}    & 2 / - / R \\ \cline{4-5} \cline{7-8} \cline{10-11} 
\multicolumn{1}{?c|}{}                          &                        & \multicolumn{1}{c|}{}                     & \multicolumn{1}{c|}{Unified} & 3 / 3 / R & \multicolumn{1}{c|}{}                     & \multicolumn{1}{c|}{Unified} & 3 / 5 / R & \multicolumn{1}{c|}{}                     & \multicolumn{1}{c|}{Unified} & 3 / 3 / R \\ \cline{4-5} \cline{7-8} \cline{10-11} 
\multicolumn{1}{?c|}{}                          &                        & \multicolumn{1}{c|}{}                     & \multicolumn{1}{c|}{MHSA}    & 2 / - / R & \multicolumn{1}{c|}{}                     & \multicolumn{1}{c|}{MHSA}    & 2 / - / R & \multicolumn{1}{c|}{}                     & \multicolumn{1}{c|}{MHSA}    & 2 / - / R \\ \cline{4-5} \cline{7-8} \cline{10-11} 
\multicolumn{1}{?c|}{}                          &                        & \multicolumn{1}{c|}{}                     & \multicolumn{1}{c|}{Fused}   & 2 / 3 / R & \multicolumn{1}{c|}{}                     & \multicolumn{1}{c|}{Unified} & 3 / 5 / R & \multicolumn{1}{c|}{}                     & \multicolumn{1}{c|}{Unified} & 3 / 3 / R \\ \cline{4-5} \cline{7-8} \cline{10-11} 
\multicolumn{1}{?c|}{}                          &                        & \multicolumn{1}{c|}{}                     & \multicolumn{1}{c|}{MHSA}    & 2 / - / R & \multicolumn{1}{c|}{}                     & \multicolumn{1}{c|}{MHSA}    & 2 / - / R & \multicolumn{1}{c|}{}                     & \multicolumn{1}{c|}{MHSA}    & 2 / - / R \\  \cline{4-5} \cline{7-8} \cline{10-11}  
\multicolumn{1}{?c|}{}                          &                        & \multicolumn{1}{c|}{}                     & \multicolumn{1}{c|}{Unified} & 4 / 3 / R & \multicolumn{1}{c|}{}                     & \multicolumn{1}{c|}{Unified} & 4 / 5 / R & \multicolumn{1}{l|}{}                     & \multicolumn{1}{c|}{Unified} & 4 / 3 / R \\  \cline{4-5} \cline{7-8} \cline{10-11}  
\multicolumn{1}{?c|}{}                          &                        & \multicolumn{1}{c|}{}                     & \multicolumn{1}{c|}{Unified} & 2 / 5 / R & \multicolumn{1}{c|}{}                     & \multicolumn{1}{c|}{Unified} & 4 / 5 / R & \multicolumn{1}{l|}{}                     & \multicolumn{1}{c|}{Unified} & 2 / 3 / R \\  \cline{4-5} \cline{7-8} \cline{10-11} 
\multicolumn{1}{?c|}{}                          &                        & \multicolumn{1}{c|}{}                     & \multicolumn{1}{c|}{MHSA}    & 2 / - / R & \multicolumn{1}{c|}{}                     & \multicolumn{1}{c|}{MHSA}    & 2 / - / R & \multicolumn{1}{l|}{}                     & \multicolumn{1}{c|}{MHSA}    & 2 / - / R \\  \cline{4-5} \cline{7-8} \cline{10-11} 
\multicolumn{1}{?c|}{}                          &                        & \multicolumn{1}{c|}{}                     & \multicolumn{1}{c|}{Unified} & 3 / 3 / R & \multicolumn{1}{c|}{}                     & \multicolumn{1}{c|}{Unified} & 3 / 5 / R & \multicolumn{1}{l|}{}                     & \multicolumn{1}{c|}{Unified} & 3 / 3 / R \\ \hline
\multicolumn{1}{?c|}{Embed 3}                   & 1/32                   & \multicolumn{1}{c|}{248}                  & \multicolumn{1}{c|}{MHSA DS} & 4 / - / R & \multicolumn{1}{c|}{224}                  & \multicolumn{1}{c|}{MHSA DS} & 4 / - / R & \multicolumn{1}{c|}{224}                  & \multicolumn{1}{c|}{MHSA DS} & 2 / - / R \\ \hline
\multicolumn{1}{?c|}{\multirow{8}{*}{Stage 4}}  & \multirow{8}{*}{1/32}  & \multicolumn{1}{c|}{\multirow{8}{*}{248}} & \multicolumn{1}{c|}{MHSA}    & 2 / - / R & \multicolumn{1}{c|}{\multirow{8}{*}{224}} & \multicolumn{1}{c|}{MHSA}    & 2 / - / G & \multicolumn{1}{c|}{\multirow{8}{*}{224}} & \multicolumn{1}{c|}{MHSA}    & 2 / - / R \\ \cline{4-5} \cline{7-8} \cline{10-11} 
\multicolumn{1}{?c|}{}                          &                        & \multicolumn{1}{c|}{}                     & \multicolumn{1}{c|}{Fused}   & 3 / 3 / R & \multicolumn{1}{c|}{}                     & \multicolumn{1}{c|}{Unified} & 3 / 5 / G & \multicolumn{1}{c|}{}                     & \multicolumn{1}{c|}{Unified} & 3 / 3 / R \\ \cline{4-5} \cline{7-8} \cline{10-11} 
\multicolumn{1}{?c|}{}                          &                        & \multicolumn{1}{c|}{}                     & \multicolumn{1}{c|}{MHSA}    & 2 / - / G & \multicolumn{1}{c|}{}                     & \multicolumn{1}{c|}{MHSA}    & 2 / - / G & \multicolumn{1}{c|}{}                     & \multicolumn{1}{c|}{MHSA}    & 2 / - / R \\ \cline{4-5} \cline{7-8} \cline{10-11} 
\multicolumn{1}{?c|}{}                          &                        & \multicolumn{1}{c|}{}                     & \multicolumn{1}{c|}{Fused}   & 3 / 3 / G & \multicolumn{1}{c|}{}                     & \multicolumn{1}{c|}{Unified} & 3 / 5 / G & \multicolumn{1}{c|}{}                     & \multicolumn{1}{c|}{Unified} & 3 / 3 / R \\ \cline{4-5} \cline{7-8} \cline{10-11} 
\multicolumn{1}{?c|}{}                          &                        & \multicolumn{1}{c|}{}                     & \multicolumn{1}{c|}{MHSA}    & 2 / - / R & \multicolumn{1}{c|}{}                     & \multicolumn{1}{c|}{MHSA}    & 2 / - / R & \multicolumn{1}{c|}{}                     & \multicolumn{1}{c|}{MHSA}    & 2 / - / R \\ \cline{4-5} \cline{7-8} \cline{10-11} 
\multicolumn{1}{?c|}{}                          &                        & \multicolumn{1}{c|}{}                     & \multicolumn{1}{c|}{Unified} & 3 / 3 / R & \multicolumn{1}{c|}{}                     & \multicolumn{1}{c|}{Unified} & 4 / 5 / R & \multicolumn{1}{c|}{}                     & \multicolumn{1}{c|}{Unified} & 3 / 3 / R \\ \cline{4-5} \cline{7-8} \cline{10-11} 
\multicolumn{1}{?c|}{}                          &                        & \multicolumn{1}{c|}{}                     & \multicolumn{1}{c|}{MHSA}    & 2 / - / G & \multicolumn{1}{c|}{}                     & \multicolumn{1}{c|}{MHSA}    & 2 / - / R & \multicolumn{1}{c|}{}                     & \multicolumn{1}{c|}{MHSA}    & 2 / - / R \\ \cline{4-5} \cline{7-8} \cline{10-11}
\multicolumn{1}{?c|}{}                          &                        & \multicolumn{1}{c|}{}                     & \multicolumn{1}{c|}{Fused}   & 3 / 3 / G & \multicolumn{1}{c|}{}                     & \multicolumn{1}{c|}{Unified} & 3 / 5 / R & \multicolumn{1}{c|}{}                     & \multicolumn{1}{c|}{Unified} & 3 / 3 / R \\ \bottomrule
\end{tabular}
\end{table*}

\begin{table*}[h]
\centering
\small
\caption{
Architecture details of searched subnets on Nano with TensorRT. We abbreviate output resolution (Res.), output channel width (Ch.), fused FFNs (Fused), Unified FFNs (unified), and Multi-head self-attention layers (MHSA), embedding layers (Embed 3), activation layers (A), GELU (G),  ReLU (R), expansion ratio (E), and kernel size (K) in the Table.
}
\label{tab:subnet_details_nano_trt}
\begin{tabular}{?cc?ccc?ccc?ccc?}
\toprule
\multicolumn{2}{?l?}{}                                                 & \multicolumn{3}{c?}{Nano\_tensorrt\_@20ms}                                           & \multicolumn{3}{c?}{Nano\_tensorrt\_@20ms\_@5M}                                      & \multicolumn{3}{c?}{Nano\_tensorrt\_@13ms}                                           \\ \toprule
\multicolumn{1}{?c|}{Stage}                    & Res.              & \multicolumn{1}{c|}{Ch.}              & \multicolumn{1}{c|}{Type}    & E / K / A & \multicolumn{1}{c|}{Ch.}              & \multicolumn{1}{c|}{Type}    & E / K / A & \multicolumn{1}{c|}{Ch.}              & \multicolumn{1}{c|}{Type}    & E / K / A \\ \hline
\multicolumn{1}{?c|}{Conv}                & 1/4                   & \multicolumn{1}{c|}{32}                   & \multicolumn{1}{c|}{Conv}    & - / 3 / R & \multicolumn{1}{c|}{32}                   & \multicolumn{1}{c|}{Conv}    & - / 3 / R & \multicolumn{1}{c|}{32}                   & \multicolumn{1}{c|}{Conv}    & - / 3 / R \\ \hline
\multicolumn{1}{?c|}{\multirow{2}{*}{Stage 1}} & \multirow{2}{*}{1/4}  & \multicolumn{1}{c|}{\multirow{2}{*}{32}}  & \multicolumn{1}{c|}{Fused}   & 3 / 3 / R & \multicolumn{1}{c|}{\multirow{2}{*}{32}}  & \multicolumn{1}{c|}{Fused}   & 4 / 3 / G & \multicolumn{1}{c|}{\multirow{2}{*}{32}}  & \multicolumn{1}{c|}{Unified} & 2 / 3 / R \\ \cline{4-5} \cline{7-8} \cline{10-11} 
\multicolumn{1}{?c|}{}                         &                       & \multicolumn{1}{c|}{}                     & \multicolumn{1}{c|}{Fused}   & 3 / 3 / R & \multicolumn{1}{c|}{}                     & \multicolumn{1}{c|}{Fused}   & 3 / 3 / R & \multicolumn{1}{c|}{}                     & \multicolumn{1}{c|}{Fused}   & 3 / 3 / R \\ \hline
\multicolumn{1}{?c|}{Embed 1}                  & 1/8                   & \multicolumn{1}{c|}{64}                   & \multicolumn{1}{c|}{Conv}    & -         & \multicolumn{1}{c|}{64}                   & \multicolumn{1}{c|}{Conv}    & -         & \multicolumn{1}{c|}{64}                   & \multicolumn{1}{c|}{Conv}    & -         \\ \hline
\multicolumn{1}{?c|}{\multirow{2}{*}{Stage 2}} & \multirow{2}{*}{1/8}  & \multicolumn{1}{c|}{\multirow{2}{*}{64}}  & \multicolumn{1}{c|}{Fused}   & 3 / 3 / G & \multicolumn{1}{c|}{\multirow{2}{*}{64}}  & \multicolumn{1}{c|}{Fused}   & 2 / 3 / G & \multicolumn{1}{c|}{\multirow{2}{*}{64}}  & \multicolumn{1}{c|}{Unified} & 2 / 3 / R \\ \cline{4-5} \cline{7-8} \cline{10-11} 
\multicolumn{1}{?c|}{}                         &                       & \multicolumn{1}{c|}{}                     & \multicolumn{1}{c|}{Fused}   & 3 / 3 / G & \multicolumn{1}{c|}{}                     & \multicolumn{1}{c|}{Fused}   & 2 / 3 / G & \multicolumn{1}{c|}{}                     & \multicolumn{1}{c|}{Unified} & 2 / 3 / R \\ \hline
\multicolumn{1}{?c|}{Embed 2}                  & 1/16                  & \multicolumn{1}{c|}{96}                   & \multicolumn{1}{c|}{Conv}    & -         & \multicolumn{1}{c|}{120}                  & \multicolumn{1}{c|}{Conv}    & -         & \multicolumn{1}{c|}{96}                   & \multicolumn{1}{c|}{Conv}    & -         \\ \hline
\multicolumn{1}{?c|}{\multirow{9}{*}{Stage 3}} & \multirow{9}{*}{1/16} & \multicolumn{1}{c|}{\multirow{9}{*}{96}}  & \multicolumn{1}{c|}{Fused}   & 3 / 3 / R & \multicolumn{1}{c|}{\multirow{8}{*}{120}} & \multicolumn{1}{c|}{Unified} & 3 / 5 / R & \multicolumn{1}{c|}{\multirow{6}{*}{96}}  & \multicolumn{1}{c|}{Fused}   & 3 / 3 / R \\ \cline{4-5} \cline{7-8} \cline{10-11} 
\multicolumn{1}{?c|}{}                         &                       & \multicolumn{1}{c|}{}                     & \multicolumn{1}{c|}{Fused}   & 3 / 3 / R & \multicolumn{1}{c|}{}                     & \multicolumn{1}{c|}{Unified} & 3 / 5 / R & \multicolumn{1}{c|}{}                     & \multicolumn{1}{c|}{Fused}   & 3 / 3 / R \\ \cline{4-5} \cline{7-8} \cline{10-11} 
\multicolumn{1}{?c|}{}                         &                       & \multicolumn{1}{c|}{}                     & \multicolumn{1}{c|}{Fused}   & 3 / 3 / R & \multicolumn{1}{c|}{}                     & \multicolumn{1}{c|}{MHSA}    & 2 / - / G & \multicolumn{1}{c|}{}                     & \multicolumn{1}{c|}{Fused}   & 3 / 3 / R \\ \cline{4-5} \cline{7-8} \cline{10-11} 
\multicolumn{1}{?c|}{}                         &                       & \multicolumn{1}{c|}{}                     & \multicolumn{1}{c|}{Fused}   & 3 / 3 / R & \multicolumn{1}{c|}{}                     & \multicolumn{1}{c|}{Unified} & 2 / 5 / G & \multicolumn{1}{c|}{}                     & \multicolumn{1}{c|}{Fused}   & 3 / 3 / R \\ \cline{4-5} \cline{7-8} \cline{10-11} 
\multicolumn{1}{?c|}{}                         &                       & \multicolumn{1}{c|}{}                     & \multicolumn{1}{c|}{Fused}   & 3 / 3 / G & \multicolumn{1}{c|}{}                     & \multicolumn{1}{c|}{Unified} & 2 / 5 / R & \multicolumn{1}{c|}{}                     & \multicolumn{1}{c|}{Fused}   & 3 / 3 / G \\ \cline{4-5} \cline{7-8} \cline{10-11} 
\multicolumn{1}{?c|}{}                         &                       & \multicolumn{1}{c|}{}                     & \multicolumn{1}{c|}{Fused}   & 3 / 3 / G & \multicolumn{1}{c|}{}                     & \multicolumn{1}{c|}{MHSA}    & 2 / - / G & \multicolumn{1}{c|}{}                     & \multicolumn{1}{c|}{Fused}   & 3 / 3 / G \\ \cline{4-5} \cline{7-11} 
\multicolumn{1}{?c|}{}                         &                       & \multicolumn{1}{c|}{}                     & \multicolumn{1}{c|}{Fused}   & 3 / 3 / R & \multicolumn{1}{c|}{}                     & \multicolumn{1}{c|}{Unified} & 3 / 5 / G & \multicolumn{3}{c?}{\multirow{3}{*}{N/A}}                                            \\ \cline{4-5} \cline{7-8}
\multicolumn{1}{?c|}{}                         &                       & \multicolumn{1}{c|}{}                     & \multicolumn{1}{c|}{Fused}   & 3 / 3 / G & \multicolumn{1}{c|}{}                     & \multicolumn{1}{c|}{Unified} & 3 / 5 / R & \multicolumn{3}{c?}{}                                                                \\ \cline{4-8}
\multicolumn{1}{?c|}{}                         &                       & \multicolumn{1}{c|}{}                     & \multicolumn{1}{c|}{Fused}   & 3 / 3 / G & \multicolumn{3}{c?}{N/A}                                                             & \multicolumn{3}{c?}{}                                                                \\ \hline
\multicolumn{1}{?c|}{Embed 3}                  & 1/32                  & \multicolumn{1}{c|}{224}                  & \multicolumn{1}{c|}{MHSA DS} & 2 / - / G & \multicolumn{1}{c|}{248}                  & \multicolumn{1}{c|}{MHSA DS} & 2 / - / R & \multicolumn{1}{c|}{224}                  & \multicolumn{1}{c|}{MHSA DS} & 2 / - / R \\ \hline
\multicolumn{1}{?c|}{\multirow{8}{*}{Stage 4}} & \multirow{8}{*}{1/32} & \multicolumn{1}{c|}{\multirow{8}{*}{224}} & \multicolumn{1}{c|}{Fused}   & 3 / 3 / G & \multicolumn{1}{c|}{\multirow{7}{*}{248}} & \multicolumn{1}{c|}{MHSA}    & 2 / - / G & \multicolumn{1}{c|}{\multirow{6}{*}{224}} & \multicolumn{1}{c|}{MHSA}    & 2 / - / R \\ \cline{4-5} \cline{7-8} \cline{10-11} 
\multicolumn{1}{?c|}{}                         &                       & \multicolumn{1}{c|}{}                     & \multicolumn{1}{c|}{MHSA}    & 2 / - / G & \multicolumn{1}{c|}{}                     & \multicolumn{1}{c|}{Unified} & 3 / 5 / G & \multicolumn{1}{c|}{}                     & \multicolumn{1}{c|}{Fused}   & 2 / 3 / R \\ \cline{4-5} \cline{7-8} \cline{10-11} 
\multicolumn{1}{?c|}{}                         &                       & \multicolumn{1}{c|}{}                     & \multicolumn{1}{c|}{Fused}   & 3 / 3 / G & \multicolumn{1}{c|}{}                     & \multicolumn{1}{c|}{MHSA}    & 2 / - / G & \multicolumn{1}{c|}{}                     & \multicolumn{1}{c|}{Fused}   & 2 / 3 / G \\ \cline{4-5} \cline{7-8} \cline{10-11} 
\multicolumn{1}{?c|}{}                         &                       & \multicolumn{1}{c|}{}                     & \multicolumn{1}{c|}{Fused}   & 3 / 3 / G & \multicolumn{1}{c|}{}                     & \multicolumn{1}{c|}{Unified} & 3 / 5 / G & \multicolumn{1}{c|}{}                     & \multicolumn{1}{c|}{MHSA}    & 2 / - / G \\ \cline{4-5} \cline{7-8} \cline{10-11} 
\multicolumn{1}{?c|}{}                         &                       & \multicolumn{1}{c|}{}                     & \multicolumn{1}{c|}{Fused}   & 3 / 3 / G & \multicolumn{1}{c|}{}                     & \multicolumn{1}{c|}{Unified} & 3 / 5 / G & \multicolumn{1}{c|}{}                     & \multicolumn{1}{c|}{Fused}   & 3 / 3 / G \\ \cline{4-5} \cline{7-8} \cline{10-11} 
\multicolumn{1}{?c|}{}                         &                       & \multicolumn{1}{c|}{}                     & \multicolumn{1}{c|}{Fused}   & 3 / 3 / R & \multicolumn{1}{c|}{}                     & \multicolumn{1}{c|}{MHSA}    & 2 / - / R & \multicolumn{1}{c|}{}                     & \multicolumn{1}{c|}{Fused}   & 2 / 3 / G \\ \cline{4-5} \cline{7-11} 
\multicolumn{1}{?c|}{}                         &                       & \multicolumn{1}{c|}{}                     & \multicolumn{1}{c|}{MHSA}    & 2 / - / G & \multicolumn{1}{c|}{}                     & \multicolumn{1}{c|}{Unified} & 3 / 5 / R & \multicolumn{3}{c?}{\multirow{2}{*}{N/A}}                                            \\ \cline{4-8}
\multicolumn{1}{?c|}{}                         &                       & \multicolumn{1}{c|}{}                     & \multicolumn{1}{c|}{Fused}   & 3 / 3 / G & \multicolumn{3}{c?}{N/A}                                                             & \multicolumn{3}{c?}{}                                                                \\ \bottomrule
\end{tabular}
\end{table*}

\end{document}